%% file: template.tex
\documentclass{article}
\PassOptionsToPackage{numbers, compress}{natbib}
\usepackage[preprint]{neurips_2025}
\usepackage[utf8]{inputenc}
\usepackage[T1]{fontenc}
\usepackage{hyperref}
\usepackage{url}
\usepackage{booktabs}
\usepackage{amsfonts}
\usepackage{nicefrac}
\usepackage{microtype}
\usepackage{wrapfig}
\usepackage{cutwin}
\usepackage{lipsum}
\usepackage{caption}
\usepackage{CJK}
\usepackage[table]{xcolor}
\usepackage{arydshln}  
\input{tool}

\title{%
TACTIC: Translation Agents with Cognitive-Theoretic Interactive Collaboration
}

\author{
Weiya Li$^{1}$\hspace{0.2cm}
Junjie Chen$^{2,}$\thanks{Junjie is a research assistant at the EPIC Lab, Shanghai Jiao Tong University, working with Linfeng Zhang.}\hspace{0.3cm}
Bei Li$^{3}$\hspace{0.2cm}
Boyang Liu$^{4}$\hspace{0.2cm}
\textbf{Zichen Wen}$^{2}$\hspace{0.2cm}
\textbf{Nuanqiao Shan}$^{5}$\hspace{0.2cm}\\
\textbf{Xiaoqian Liu}$^{2,6}$\hspace{0.2cm}
\textbf{Anping Liu}$^{1}$\hspace{0.2cm}
\textbf{Huajie Liu}$^{1}$\hspace{0.2cm}
\textbf{Hu Song}$^{1}$\hspace{0.2cm}
\textbf{Linfeng Zhang}$^{2,}$\thanks{Corresponding author.}
\vspace{1.5mm}\\
$^{1}$ Big Data\&AI Lab, ICBC\hspace{0.5cm}
$^{2}$ Shanghai Jiao Tong University\hspace{0.5cm}
$^{3}$ Meituan Inc.
\vspace{0.5mm}\\
$^{4}$ Tongji University\hspace{0.5cm}
$^{5}$ Fudan University\hspace{0.5cm}
$^{6}$ Northeastern University
\vspace{1.5mm}\\
{\tt weiyali126@outlook.com\hspace{0.5cm}jorji.chen@gmail.com}\\
{\tt libei17@meituan.com\hspace{0.5cm}zhanglinfeng@sjtu.edu.cn}\\
}

\begin{document}
\maketitle

\renewcommand{\thefootnote}{1}
\footnotetext{The full name of this institution is the Big Data\&Artificial Intelligence Laboratory, Industrial and Commercial Bank of China.}
\renewcommand{\thefootnote}{\arabic{footnote}}

\thispagestyle{plain}
\begin{abstract}
\input{abstract}
\end{abstract}
\input{main}
\end{document}

%% file: tool.tex
\usepackage{enumitem}
\usepackage{verbatim}
\usepackage{xspace}
\usepackage{diagbox}
\usepackage{array}
\usepackage{cite}

\usepackage{multirow}
\usepackage{makecell}
\usepackage{diagbox}
\usepackage{url}
\usepackage{comment}
\usepackage{booktabs}
\usepackage{amsmath}
\usepackage{scalerel}
\usepackage[figuresright]{rotating}
\usepackage{threeparttable}
\usepackage{algorithm}
\usepackage{algorithmic}
\usepackage{siunitx}
\usepackage{stackengine}
\usepackage{pifont}
\usepackage{tabularx}
\usepackage{listings}

\usepackage{tcolorbox}
\definecolor{grey}{RGB}{128,128,128}

\usepackage{mathtools}
\usepackage{multirow}
\usepackage{multicol}
\usepackage{booktabs}
\usepackage{subfigure}

\usepackage{amsmath,amsfonts,bm}

\def\Figref#1{Figure~\ref{#1}}

\def\eqref#1{equation~\ref{#1}}

\def\Algref#1{Algorithm~\ref{#1}}

\def\1{\bm{1}}

\DeclareMathAlphabet{\mathsfit}{\encodingdefault}{\sfdefault}{m}{sl}
\SetMathAlphabet{\mathsfit}{bold}{\encodingdefault}{\sfdefault}{bx}{n}

\definecolor{mygreen}{rgb}{0.7, 1.0, 0.7}
\definecolor{myblue}{rgb}{0.7, 0.8, 1.0}
\newcommand{\percentbarri}[4]{%
  \rlap{\color{#3}\hspace*{-3pt}\rule{#2 em}{1.6ex}}%
  \makebox[#4 em][r]{#1}%
}

\definecolor{promptcolor}{RGB}{90,90,160} 

%% file: abstract.tex
Machine translation has long been a central task in natural language processing.
With the rapid advancement of large language models (LLMs), there has been remarkable progress in translation quality, including strong performance in zero-shot and few-shot settings.
However, fully realizing the translation potential of LLMs remains an open challenge.
Recent studies have explored multi-agent systems to decompose complex translation tasks into collaborative subtasks, showing initial promise in enhancing translation quality through agent cooperation and specialization.
Nevertheless, existing multi-agent translation frameworks largely neglect foundational insights from cognitive translation studies.
These insights emphasize how human translators employ different cognitive strategies, such as balancing literal and free translation, refining expressions based on context, and iteratively evaluating outputs.
To address this limitation, we propose a cognitively informed multi-agent framework called TACTIC, which stands for \underline{T}ranslation \underline{A}gents with \underline{C}ognitive-\underline{T}heoretic \underline{I}nteractive \underline{C}ollaboration.
The framework comprises six functionally distinct agents that mirror key cognitive processes observed in human translation behavior.
These include agents for drafting, refinement, evaluation, scoring, context reasoning, and external knowledge gathering. By simulating an interactive and theory-grounded translation workflow, TACTIC effectively leverages the full capacity of LLMs for high-quality translation. Experimental results on diverse language pairs from the FLORES-200 and WMT24 benchmarks show that our method consistently achieves state-of-the-art performance. Using DeepSeek-V3 as the base model, TACTIC surpasses GPT-4.1 by an average of +0.6 XCOMET and +1.18 COMETKIWI-23. Compared to DeepSeek-R1, it further improves by +0.84 XCOMET and +2.99 COMETKIWI-23. Code is available at https://github.com/weiyali126/TACTIC.

%% file: main.tex
\section{Introduction}

Machine Translation has been approached as a conditional sequence generation task, with the goal of learning a function that maps text from a source language to a target language \citep{bahdanau2014neural, sutskever2014sequence,vaswani2017attention,wang-etal-2019-learning}. Recently, large language models (LLMs) have demonstrated strong performance in translation~\citep{hendy2023GPTMT,ALMA2024xu,lu2024llamax}. Unlike conventional neural machine translation (NMT) systems, LLMs operate as general-purpose sequence predictors, leveraging prompt-based conditioning and broad pretraining to perform translation in a manner more akin to human translators. This paradigm shift prompts a fundamental question: \textit{what are the most essential elements in the human translation process?}

While this question remains unresolved, advancements in cognitive science and cutting-edge technologies have continually deepened our understanding, which is reflected in the evolution of state-of-the-art translation methods. Just as NMT surpassed traditional machine translation by emulating human cognitive processes through deep learning and contextual understanding, LLMs have further advanced translation technology. Building upon NMT, LLMs capture nuances of human translation, including context comprehension, diverse translation strategies, rich linguistic knowledge, and cross-task adaptability, thereby outperforming traditional NMT systems and marking a significant shift in the machine translation landscape~\citep{deutsch2025wmt24++}.

\textbf{But can we go even further?} Is there a principled way to model the human translation process more faithfully---beyond what current LLMs can do? \textit{Cognitive Translation Studies} (CTS)~\citep{halverson2010cognitive} offers a compelling framework in this regard. As an application of cognitive science in the field of translation, CTS focuses on the cognitive elements involved in the translation process---such as the nature, mechanisms, and stages. Its primary goal is to understand the underlying cognitive processes behind ``cognition'' within the context of translation~\citep{cortese1999cognitive}. To date, CTS has evolved into a comprehensive discipline encompassing a wide range of foundational perspectives~\citep{pacte2008building,schwieter2020current,yi2020overview,Yuhan_Zhylin_Antonyuk-Kyrychenko_Popova_Harbar_2024}, in this work, we focus on three core concepts: \textbf{cognitive strategies}~\citep{kussmaul1995training}, \textbf{cognitive processing}~\citep{gile2009basic}, and \textbf{contextual cognition}~\citep{risku2010cognitive}.

\begin{itemize}[leftmargin=*]
\item \textbf{Cognitive Strategies} refer to the varied approaches human translators adopt, \emph{e.g.,} literal translation, paraphrasing, and free adaptation, based on communicative intent and contextual demands.
\item \textbf{Cognitive Processing} concerns the mental operations involved in translation, including comprehension, memory access, and linguistic reformulation.
\item \textbf{Contextual Cognition} captures how translators integrate prior knowledge and discourse context to produce accurate and coherent translations.
\end{itemize}

Motivated by this cognitive perspective, we propose \textbf{TACTIC} (\underline{T}ranslation \underline{A}gents with \underline{C}ognitive-\underline{T}heoretic \underline{I}nteractive \underline{C}ollaboration) --- a multi-agent translation framework that explicitly aligns with the core components of CTS. It integrates diverse translation strategies, linguistic evaluation principles, and contextual reasoning into a unified multi-agent architecture. Specifically, TACTIC comprises six agents, each emulating specific cognitive functions in the human translation process: \textit{ResearchAgent}, \textit{ContextAgent}, \textit{DraftAgent}, \textit{RefinementAgent}, \textit{EvaluationAgent}, and \textit{ScoreAgent}. These agents operate in two stages: an initial fast translation phase and an iterative refinement phase. By dynamically adjusting the translation pathway, TACTIC simulates the iterative and context-aware nature of human translation cognition.

By aligning each agent with specific dimensions of cognitive translation theory, TACTIC operationalizes CTS into a structured computational paradigm. This framework not only yields empirical improvements in translation quality but also offers theoretical interpretability grounded in cognitive science. 
Through TACTIC, we aim to bridge the gap between human translation cognition and machine translation architectures, providing a novel perspective for translation modeling in the era of large language models.Grounded in this cognitive perspective, our work makes three key contributions:

\begin{itemize}[leftmargin=*]
\item \textbf{Cognitive-Inspired Multi-Agent Framework:} We propose TACTIC, a multi-agent translation architecture explicitly aligned with the core components of Cognitive Translation Studies. Each agent simulates a specific cognitive function in human translation, forming an interpretable system.

\item \textbf{Operationalizing Translation Strategies and Processing:} We models strategic variation through the \textit{DraftAgent}, which generates literal, sense-for-sense, and free translations. The \textit{EvaluationAgent} and \textit{ScoreAgent} simulate cognitive processing by conducting feedback-based assessment and scoring grounded in the classical \textit{faithfulness}, \textit{expressiveness}, and \textit{elegance} framework.

\item \textbf{Contextual Cognition:} We implement the \textit{ContextAgent} to infer and expand the discourse context of the source text, incorporating stylistic, situational, and audience-related cues to enhance contextual awareness and coherence, aligning with the contextual cognition dimension of CTS.
\end{itemize}

\section{TACTIC}

\begin{figure}
\centering
\includegraphics[width=\linewidth]{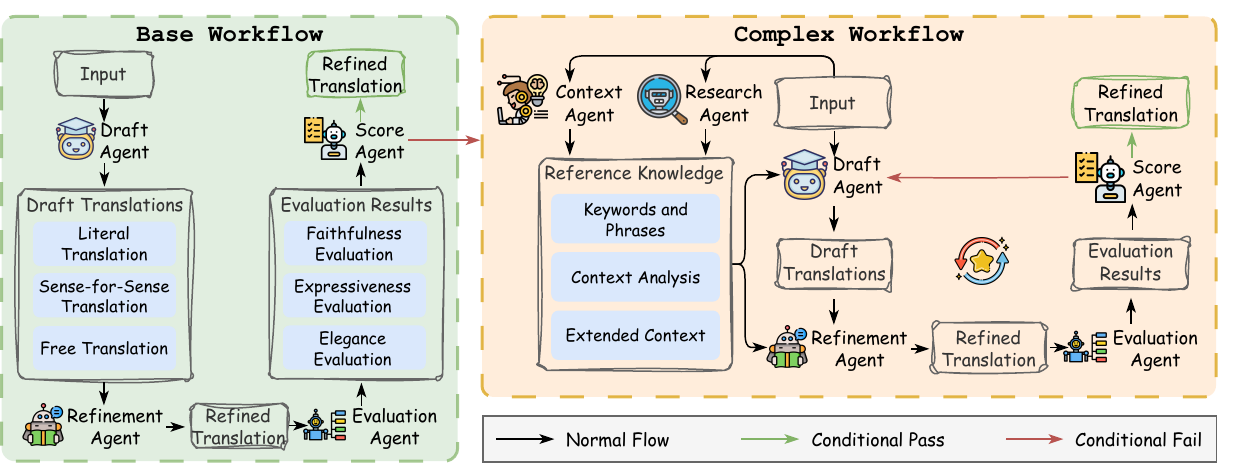}
\caption{
Overall agent collaboration workflow in the TACTIC framework. The figure depicts a cognitively inspired, modular system comprising six specialized agents: \ding{172}~\textit{DraftAgent} explores stylistic diversity through multiple translation strategies; \ding{173}~\textit{RefinementAgent} consolidates these drafts into a coherent output; \ding{174}~\textit{EvaluationAgent} applies multidimensional quality checks; \ding{175}~\textit{ScoreAgent} decides whether the result meets quality expectations. When challenges arise, the system escalates to a more reflective mode: \ding{176}~\textit{ResearchAgent} and \ding{177}~\textit{ContextAgent} inject additional knowledge and situational awareness to support iterative improvement.
This dual-layered workflow reflects the flexible reasoning patterns of human translators, who shift between fluent execution and deeper analytical engagement depending on task complexity.
}
\label{fig:tactic-framework}
\end{figure}

Inspired by CTS, we introduce TACTIC, a modular agent-based framework designed to enhance translation quality through cognitively informed collaboration among six specialized agents. An overview of the framework is illustrated in \Figref{fig:tactic-framework}.
Specifically, \ding{172} the \textit{DraftAgent} applies cognitive translation strategies to generate multi-style translations, including literal, sense-for-sense, and free renditions;
\ding{173} the \textit{RefinementAgent} synthesizes these multiple drafts to produce a refined translation;
\ding{174} the \textit{EvaluationAgent} assesses translations based on the cognitive dimensions of faithfulness, expressiveness, and elegance;
\ding{175} the \textit{ScoreAgent} assigns a quantitative score to the refined translation based on evaluation results;
\ding{176} the \textit{ResearchAgent} identifies related keywords and phrases informed by cognitive contextualization theories;
and \ding{177} the \textit{ContextAgent} supplements the translation process with broader contextual information to guide and enhance agent collaboration.

\begin{wraptable}{r}{0.42\textwidth}
  \captionsetup{width=0.42\textwidth, justification=raggedright, singlelinecheck=false}
  \centering
  \footnotesize
  \caption{Mapping Between CTS Concepts and TACTIC Agents.}
  \label{tab:mapping between cts and agents}
  \begin{tabularx}{0.42\textwidth}{lX}
    \toprule
    \textbf{CTS Concepts} & \textbf{TACTIC Agents} \\
    \midrule
    Contextual Cognition & ResearchAgent \\
    Contextual Cognition & ContextAgent \\
    Cognitive Strategies & DraftAgent \\
    Cognitive Processing & RefinementAgent \\
    Cognitive Processing & EvaluationAgent \\
    Cognitive Processing & ScoreAgent \\
    \bottomrule
  \end{tabularx}
\end{wraptable}

The collaboration among these agents is structured into a coherent workflow.
Initially, the system operates under a \textit{base workflow}: \textit{DraftAgent} generates multiple translation styles, \textit{RefinementAgent} integrates them into a single refined translation, \textit{EvaluationAgent} provides a multidimensional evaluation, and \textit{ScoreAgent} determines whether the translation meets a pre-defined quality threshold.
If the threshold is achieved, the refined translation is accepted as the final output.
However, if the translation quality falls short, we are motivated to introduce a \textit{complex workflow}, wherein \textit{ResearchAgent} and \textit{ContextAgent} are activated to gather domain-relevant lexical and contextual resources. These resources are then fed back into the drafting and refining stages, enabling iterative improvement until the translation meets the desired standards.

\subsection{Agent Design}

The design of each agent in the TACTIC framework is grounded in core principles from cognitive translation studies~\citep{halverson2010cognitive} , including cognitive translation strategies~\citep{kussmaul1995training}, cognitive processing models~\citep{gile2009basic}, and context-based cognitive theories~\citep{risku2010cognitive}, as illustrated in Table~\ref{tab:mapping between cts and agents}. These theoretical underpinnings guide the functional division and interaction protocols among agents, enabling a cognitively informed, modular architecture that mirrors human translation behavior.
We introduce the detailed design of each agent as follows:

\begin{itemize}[leftmargin=*]

\item \raisebox{-.4\baselineskip}{\includegraphics[height=1.3\baselineskip]{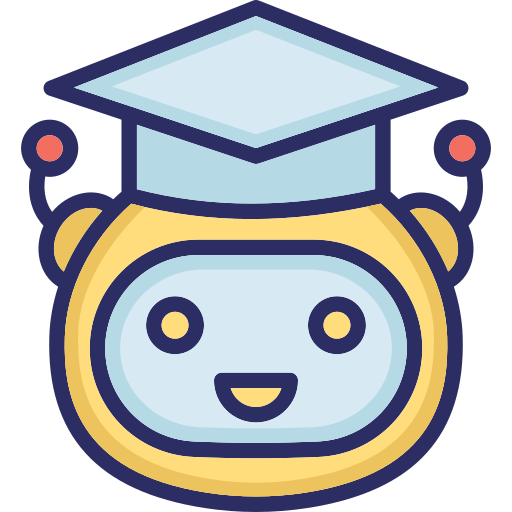}}~~\textbf{\textit{DraftAgent.}}
The \textit{DraftAgent} is inspired by the cognitive strategy theory in translation, particularly the notion that translators often engage in multiple stylistic pathways (e.g., literal, sense-for-sense, and free translation) depending on task demands and textual properties. Accordingly, this agent employs a multi-style generation mechanism to produce three distinct translation variants. These styles aim to simulate the cognitive process of divergent thinking during the initial translation phase, ensuring broad semantic and stylistic coverage.

\item \raisebox{-.4\baselineskip}{\includegraphics[height=1.3\baselineskip]{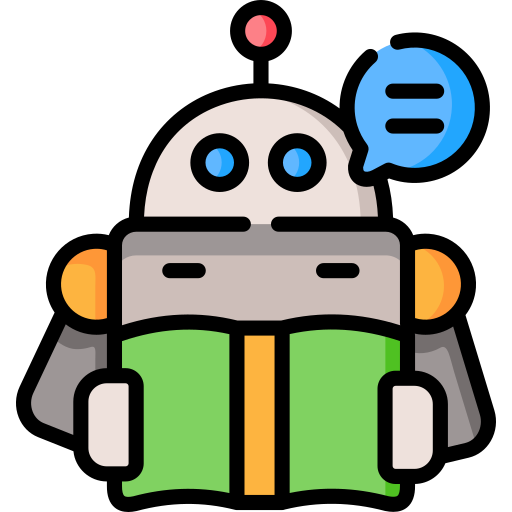}}~~\textbf{\textit{RefinementAgent.}}
The \textit{RefinementAgent} synthesizes the drafts from \textit{DraftAgent} into a single translation that is more coherent and polished. Rather than selecting the best candidate, it draws on complementary strengths across drafts to improve semantic alignment and stylistic fluency. 

\item \raisebox{-.4\baselineskip}{\includegraphics[height=1.3\baselineskip]{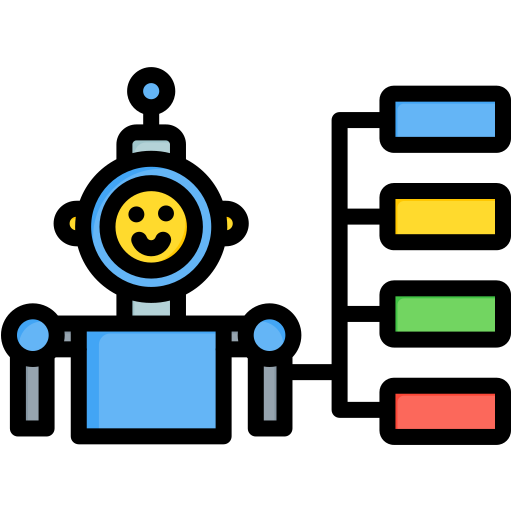}}~~\textbf{\textit{EvaluationAgent.}}
The \textit{EvaluationAgent} is grounded in cognitive models of translation processing, with particular emphasis on internal quality control and metacognitive monitoring as observed in expert translators. It systematically evaluates the refined translation along three cognitively motivated dimensions: \textit{faithfulness (semantic accuracy), expressiveness (pragmatic adequacy), and elegance (stylistic naturalness)}.
These dimensions are widely recognized in translation studies as essential to capturing both propositional meaning and communicative intent, as well as maintaining stylistic coherence and fluency. 

\item \raisebox{-.4\baselineskip}{\includegraphics[height=1.3\baselineskip]{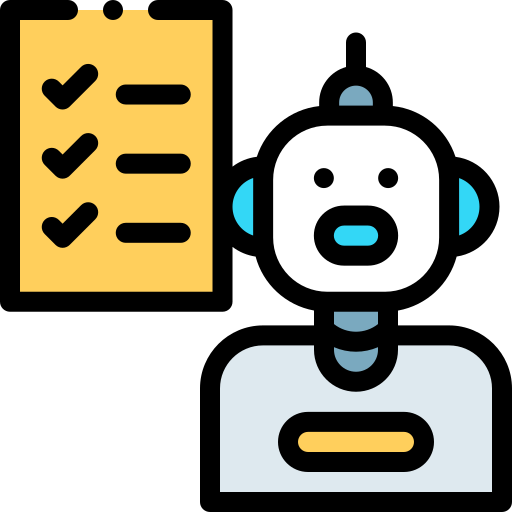}}~~\textbf{\textit{ScoreAgent.}}
Acting as a decision-making filter, the \textit{ScoreAgent} transforms qualitative assessments into a quantitative score that determines whether the refined translation meets a pre-defined threshold. This agent simulates the cognitive operation of performance monitoring, commonly discussed in psycholinguistic models of language production.

\item \raisebox{-.4\baselineskip}{\includegraphics[height=1.3\baselineskip]{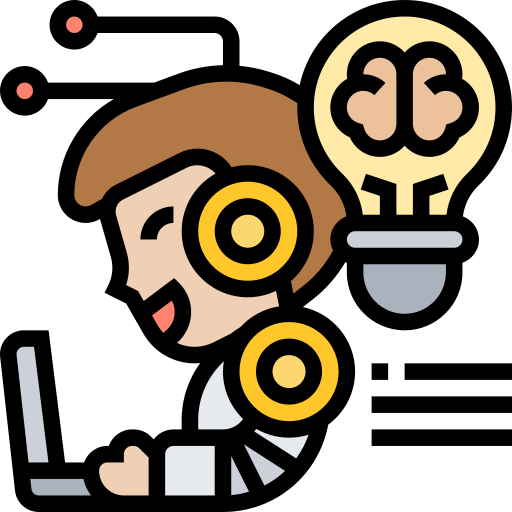}}~~\textbf{\textit{ContextAgent.}}
The \textit{ContextAgent} is based on the theory of contextual cognition in translation, which emphasizes the role of extralinguistic knowledge and situation models. This agent enriches the translation process with relevant contextual information, including domain knowledge, register, discourse intent, the preceding and following context segments expanded by the LLMs and others, all of which are critical for ensuring pragmatic adequacy and stylistic consistency in complex translation tasks.

\item \raisebox{-.4\baselineskip}{\includegraphics[height=1.3\baselineskip]{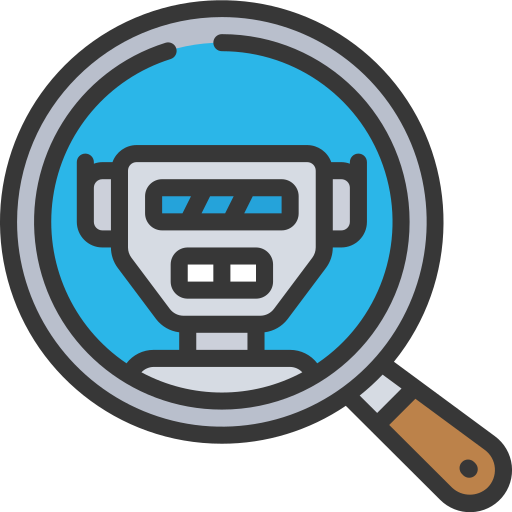}}~~\textbf{\textit{ResearchAgent.}}
Complementing the \textit{ContextAgent}, the \textit{ResearchAgent} focuses on lexical and conceptual elaboration. Informed by the contextual enrichment theory, this agent identifies relevant keywords, collocations, and conceptual associations that may support more informed and accurate translation choices. It functions as a dynamic external memory module, helping simulate the translator’s reference-gathering behavior.
\end{itemize}




\subsection{Agent Workflow}

The agent workflow in TACTIC is designed not merely as a linear processing pipeline, but as a cognitively inspired collaboration protocol that mirrors how human translators dynamically allocate effort and adapt strategies based on task complexity. As illustrated in \Figref{fig:tactic-framework} and formalized in \Algref{alg:tactic}, the workflow unfolds in two distinct layers: a \textit{Base Workflow} for routine cases, and a contextually enhanced \textit{Complex Workflow} for cognitively demanding translation tasks.
Let $x$ denote the input source text and $\tau$ a predefined quality threshold. The final output $T^*$ is produced through iterative reasoning and quality control across agents, each functioning as a specialized cognitive module.

\subsubsection{Base Workflow}

In the base workflow, translation begins with stylistic divergence.
The \textit{DraftAgent} generates three translation drafts, denoted as $T_1$, $T_2$, and $T_3$, which correspond to distinct cognitive strategies observed in human translators: \textit{literal translation}, \textit{sense-for-sense rendering}, and \textit{free adaptation}.
These stylistic variants emulate the divergent thinking phase in translation cognition, where multiple interpretive pathways are explored in parallel before convergence.

The \textit{RefinementAgent} then performs convergence by synthesizing these drafts into a refined candidate $T_r$, guided by internal heuristics for semantic cohesion and stylistic harmony. This output undergoes a triadic evaluation via the \textit{EvaluationAgent}, which scores the translation along three core axes rooted in Chinese and cognitive translation theory: faithfulness ($f$), expressiveness ($e$), and elegance ($a$). The composite score $s = \text{ScoreAgent}(f, e, a)$ quantifies the overall acceptability of the translation. If $s \geq \tau$, the workflow terminates with $T^* = T_r$.

\begin{wrapfigure}{r}{0.5\textwidth}
\begin{minipage}{\linewidth}
\vspace{-2em}
\small
\begin{algorithm}[H]
\caption{TACTIC Translation Workflow}
\begin{algorithmic}[1]
\STATE \textbf{Input:} source text $x$, quality threshold $\tau$
\STATE \textbf{Output:} refined translation $T^*$

\STATE $\{T_1, T_2, T_3\} \gets \text{DraftAgent}(x)$
\STATE $T_r \gets \text{RefinementAgent}(\{T_1, T_2, T_3\})$
\STATE $(f, e, a) \gets \text{EvaluationAgent}(T_r)$
\STATE $s \gets \text{ScoreAgent}(f, e, a)$

\IF{$s \geq \tau$}
\STATE $T^* \gets T_r$
\ELSE
\REPEAT
    \STATE $K \gets \text{ResearchAgent}(x)$
    \STATE $C \gets \text{ContextAgent}(x)$
    \STATE $\{T^{\prime}_1, T^{\prime}_2, T^{\prime}_3\} \gets \text{DraftAgent}(x; K, C)$
    \STATE $T^{\prime}_r \gets \text{RefinementAgent}(\{T^{\prime}_1, T^{\prime}_2, T^{\prime}_3\})$
    \STATE $(f^{\prime}, e^{\prime}, a^{\prime}) \gets \text{EvaluationAgent}(T^{\prime}_r)$
    \STATE $s^{\prime} \gets \text{ScoreAgent}(f^{\prime}, e^{\prime}, a^{\prime})$
\UNTIL{$s^{\prime} \geq \tau$}
\STATE $T^* \gets T^{\prime}_r$
\ENDIF
\RETURN $T^*$
\end{algorithmic}
\label{alg:tactic}
\end{algorithm}
\vspace{-0.6cm}
\end{minipage}
\end{wrapfigure}

This base process is designed to be both efficient and sufficient for straightforward translation tasks, particularly in cases where the source meaning is explicit, syntactic complexity is low, and stylistic or pragmatic demands are moderate.
It provides a reliable pipeline for generating high-quality translations without the need for external context or domain-specific adaptation, making it well-suited for general-purpose or low-stakes scenarios.
\subsubsection{Complex Workflow}
When the base workflow fails to meet the predefined quality threshold ($s < \tau$), TACTIC transitions into the complex workflow, which simulates an advanced cognitive loop inspired by expert-level translation behavior. This design reflects the notion that professional translators do not simply refine surface-level variations, but instead actively seek new knowledge, reassess the communicative context, and explore alternative interpretations when initial solutions prove inadequate.

At each iteration of this process, the system invokes two auxiliary agents: the \textit{ResearchAgent} and the \textit{ContextAgent}. The \textit{ResearchAgent} extracts relevant domain-specific keywords, technical terms, idiomatic expressions, and collocations ($K$) that may enhance lexical precision and terminological adequacy. Simultaneously, the \textit{ContextAgent} retrieves contextual parameters ($C$) that characterize the communicative situation, including but not limited to the speaker’s intent, discourse framing, target audience, register, the preceding and following context segments expanded by the LLMs. Together, these agents provide an enriched cognitive grounding that supports deeper semantic interpretation and stylistic alignment.
Using the dynamically updated $(K, C)$ pair, the \textit{DraftAgent} generates a new triad of translation candidates $\{T^{\prime}_1, T^{\prime}_2, T^{\prime}_3\}$, each representing a different cognitive translation strategy adapted to the enriched context. These are then synthesized by the \textit{RefinementAgent} into a unified translation $T^{\prime}_r$, which is subsequently assessed along the three core dimensions of faithfulness ($f^{\prime}$), expressiveness ($e^{\prime}$), and elegance ($a^{\prime}$) by the \textit{EvaluationAgent}. The composite quality score $s^{\prime} = \text{ScoreAgent}(f^{\prime}, e^{\prime}, a^{\prime})$ determines whether the iteration has reached the quality threshold.
If $s^{\prime} \geq \tau$, the workflow terminates with $T^* = T^{\prime}_r$ as the final translation. Otherwise, the system enters another iteration, during which the \textit{ResearchAgent} and \textit{ContextAgent} are re-invoked to update $(K, C)$ based on the latest interpretive needs. The recalibration loop terminates once the score threshold or stopping conditions are satisfied.

This iterative, cognitively augmented loop allows TACTIC to engage in a form of reflective problem-solving, dynamically adjusting its interpretive scope and stylistic output based on both internal evaluation and additional informational feedback. It mirrors the human expert translator's capacity for adaptive reasoning and context-sensitive decision-making, particularly in situations involving domain complexity, pragmatic ambiguity, or high stylistic demands.

\section{Experiments}

\subsection{Experimental Settings}

\paragraph{Models.}
We employ a set of mainstream models as our backend models and baselines. The non-reasoning models include various sizes of Qwen2.5 series~\citep{yang2024qwen2}(7B, 14B, 32B, 72B), Deepseek-V3-0324~\citep{liu2024deepseek}, and OpenAI's GPT-4.1~\citep{openai2025gpt41}. The reasoning-capable models include QwQ-32B~\citep{qwen2024qwq32b} and Deepseek-R1-0120~\citep{guo2025deepseek}. For non-reasoning models, we evaluate under both zero-shot and few-shot settings (using five examples sourced from~\citep{xu2024a}, these examples are also utilized in our non-reasoning \textit{DraftAgent} and \textit{RefinementAgent}). For reasoning-capable models, we adopt only the zero-shot setting. 

\vspace{-0.8em}
\paragraph{Datasets.}
We evaluate our framework using two benchmark datasets: FLORES-200~\citep{costa2022no} and WMT24~\citep{kocmi-etal-2024-findings}. Our focus is on English-centric translation tasks, encompassing both English-to-X (en$\rightarrow$xx) and X-to-English (xx$\rightarrow$en) directions. For each dataset, we assess five language pairs: German (de), Japanese (ja), Russian (ru), Ukrainian (uk), and Chinese (zh). Since the WMT24 dataset provide only the en$\rightarrow$xx direction, we supplement the evaluation by translating from these languages into English to analyze performance in the reverse direction. All datasets are standardized using the Tower~\citep{alves2024tower} framework.

\vspace{-0.8em}
\paragraph{Evaluation Metrics.}
We focus on three primary metrics: XCOMET-XXL~\citep{guerreiro2024xcomet} and MetricX-24-XXL~\citep{juraska2024metricx24googlesubmissionwmt}, both reference-based, and COMETKIWI-23-XXL~\citep{rei2023scaling}, which is reference-free. These metrics have demonstrated high correlation with human judgments. Notably, XCOMET and MetricX were also employed in the WMT24 automatic evaluations, and we adopt their XXL variants to further enhance alignment with human assessments. In addition to these model-based metrics, we also report results on ChrF~\citep{popovic2015chrf} and sacreBLEU~\citep{post2018call} to provide a comprehensive comparison. Due to space limitations, detailed results for these supplementary metrics are presented in Appendix~\ref{Additional Results}, Table \ref{tab:main results chrf bleu metricx}.

\input{Analysis/main_result}

\subsection{Main Results}
Table~\ref{tab:main results} presents the translation performance of our proposed TACTIC methods compared to baselines in both zero-shot and few-shot settings. All evaluations were conducted rigorously using both reference-based (XCOMET) and reference-free (COMETKIWI-23) metrics to ensure comprehensive and robust conclusions. Our results demonstrate that TACTIC consistently yields substantial improvements when integrated with the same underlying translation models and agent configurations. For instance, in the few-shot setting with DeepSeek-V3, TACTIC elevates the XCOMET score from 94.90 to 96.19 and improves the COMETKIWI-23 score from 90.25 to 92.64.

Our TACTIC framework, particularly when utilizing powerful models for both the agent and translation tasks, achieves state-of-the-art or highly competitive results. The configuration employing DeepSeek-V3 for both roles (TACTIC, DeepSeek-V3 // DeepSeek-V3) registers the highest XCOMET scores on FLORES-200 en$\rightarrow$xx (\textbf{96.19}), FLORES-200 xx$\rightarrow$en (\textbf{96.69}), and WMT24 xx$\rightarrow$en (\textbf{89.07}). It also achieves a strong COMETKIWI-23 of 90.15 for FLORES-200 xx$\rightarrow$en. These results often surpass strong zero-shot baselines like GPT-4.1 (FLORES-200 xx$\rightarrow$en: XCOMET \underline{96.50}, COMETKIWI-23 88.61) and DeepSeek-R1 (WMT24 xx$\rightarrow$en: XCOMET 87.98, COMETKIWI-23 75.01) in several categories, particularly in XCOMET scores.

The results also indicate a general trend of improved performance with larger model sizes within the same family (e.g., Qwen2.5-Instruct series) for both baseline and TACTIC settings, although TACTIC consistently provides an additional uplift. The improvements brought by TACTIC are evident across both FLORES-200 and the more challenging WMT24 dataset, and in both en$\rightarrow$xx and xx$\rightarrow$en translation directions, showing the robustness and broad applicability of our approach.

\subsection{Ablation Studies}
\begin{wraptable}{r}{0.504\textwidth} 
\vspace{-1em}
\scriptsize
\centering
\caption{
Incremental evaluation of agent components in the TACTIC framework.
Each setting adds cognitively inspired modules, revealing steady improvements across metrics.
}
\vspace{-0.5cm}
\label{tab:efficiency-comparison}
\vspace{1em}
\begin{tabularx}{\linewidth}{
@{}l l l@{}
}
\textbf{Method} & \textbf{XCOMET} &  \textbf{KIWI-23} \\
\midrule
\rowcolor{gray!30}
\multicolumn{3}{c}{\textbf{\textit{en-xx}}} \\
\specialrule{0em}{1pt}{1pt}
\textbf{Zero-Shot} & \percentbarri{93.30}{3.8}{mygreen}{3} & \percentbarri{88.02}{3.8}{myblue}{3}  \\
+ \textit{Iterative Evaluation} & \percentbarri{94.35}{4.3}{mygreen}{3} & \percentbarri{89.43}{4.3}{myblue}{3}  \\
\textbf{Few-Shot} & \percentbarri{93.45}{3.9}{mygreen}{3} & \percentbarri{88.17}{3.9}{myblue}{3}  \\
+ \textit{Iterative Evaluation} & \percentbarri{94.37}{4.3}{mygreen}{3} & \percentbarri{89.29}{4.3}{myblue}{3}  \\
\specialrule{0em}{1pt}{1pt}
\hdashline
\specialrule{0em}{1pt}{1pt}
\textbf{Drafting-then-Refining} & \percentbarri{94.32}{4.3}{mygreen}{3} & \percentbarri{88.96}{4.3}{myblue}{3}  \\
+ \textit{Iterative Evaluation} & \percentbarri{94.42}{4.6}{mygreen}{3} & \percentbarri{89.16}{4.6}{myblue}{3}  \\
++ \textit{Keyword, Phrase and Context Mining}  & \percentbarri{94.53}{4.6}{mygreen}{3} & \percentbarri{89.27}{4.6}{myblue}{3}  \\
\midrule
\rowcolor{gray!30}
\multicolumn{3}{c}{\textbf{\textit{xx-en}}} \\
\specialrule{0em}{1pt}{1pt}
\textbf{Zero-Shot} & \percentbarri{94.00}{4}{mygreen}{3} & \percentbarri{86.94}{3.5}{myblue}{3}  \\
+ \textit{Iterative Evaluation} & \percentbarri{94.01}{4}{mygreen}{3} & \percentbarri{86.73}{3.4}{myblue}{3}  \\
\textbf{Few-Shot} & \percentbarri{94.05}{4.1}{mygreen}{3} & \percentbarri{86.95}{3.5}{myblue}{3}  \\
+ \textit{Iterative Evaluation} & \percentbarri{94.18}{4.3}{mygreen}{3} & \percentbarri{86.72}{3.4}{myblue}{3}  \\
\specialrule{0em}{1pt}{1pt}
\hdashline
\specialrule{0em}{1pt}{1pt}
\textbf{Drafting-then-Refining} & \percentbarri{94.24}{4.5}{mygreen}{3} & \percentbarri{87.11}{4}{myblue}{3}  \\
+ \textit{Iterative Evaluation} & \percentbarri{94.27}{4.6}{mygreen}{3} & \percentbarri{87.11}{4}{myblue}{3}  \\
++ \textit{Keyword, Phrase and Context Mining}  & \percentbarri{94.40}{4.8}{mygreen}{3} & \percentbarri{87.06}{4}{myblue}{3}
\end{tabularx}
\end{wraptable}
To assess the effectiveness of each cognitively inspired agent in TACTIC, we conducted a set of ablation studies as shown in Table ~\ref{tab:efficiency-comparison}. We can observe a clear pattern of improvement as cognitively inspired modules are added. Starting from a Zero-Shot baseline, the inclusion of the \textit{EvaluationAgent} and \textit{ScoreAgent} under the Iterative Evaluation setup results in noticeable quality gains. This supports the view that translation benefits from structured self-assessment, a hallmark of expert human translators.
Further improvements appear when the \textit{DraftAgent} and \textit{RefinementAgent} are introduced through a Drafting-then-Refining process. This reflects the widely studied two-phase model of human translation: an initial draft generation followed by semantic and stylistic adjustment.
Finally, the integration of the \textit{ResearchAgent} and \textit{ContextAgent}, which provide keyword extraction, phrase-level cues, and discourse-level context, leads to the best observed performance. These enhancements simulate how human translators activate external knowledge and contextual frames when handling complex input, especially in low-resource or ambiguous scenarios.
In summary, the results validate our design: each agent meaningfully contributes to translation quality in ways that align with cognitive theory. 

\input{Analysis/case_study_1}

\section{Analysis}

\input{Analysis/ana1}
\section{Related Work}

The remarkable progress in large language models has fundamentally transformed the landscape of machine translation, with LLM-based systems now consistently outperforming traditional NMT models~\citep{deutsch2025wmt24++}. This paradigm shift has catalyzed a growing body of research, spanning both academic inquiry and open-source innovation, into how to best harness LLMs for translation. A predominant research direction centers on developing LLMs fine-tuned for translation~\citep{xu2024x,alves2024tower,guo2025redefining}. Complementing this line of work are reasoning-enhanced LLMs~\citep{he2025r1,feng2025mt,wang2025deep}, which integrate reasoning capabilities to improve translation performance.

Recently, inspired by the rise of AI agents~\citep{li2023camel}, researchers have begun exploring multi-agent translation frameworks, where distinct agents handle various roles-such as drafting, evaluation, and refinement in a manner that mirrors collaborative human workflows, leading to significant gains in translation quality.
\citet{briakou2024translatingstepbystepdecomposingtranslation} introduced a pipeline translation framework comprising research, drafting, refinement, and proofreading phases to incrementally enhance translation quality. However, due to its inherently sequential structure, it remains challenging to guarantee that all generated translations satisfy the desired quality criteria at the point of output.
\citet{wang2025drtdeepreasoningtranslation} proposed an iterative multi-agent translation framework involving a translator, an advisor, and an evaluator. By synthesizing translation process data through this framework and training a specialized inference model, they achieved significant advances in literary translation.  A similar translation process is also explored in~\citep{feng2024tearimprovingllmbasedmachine}.
\citet{he-etal-2024-exploring} introduced a human-like translation framework in which LLMs first identify keywords, topics, and relevant examples from the source text. Candidate translations are then generated conditioned on this information and ranked using external quality estimation (QE) methods to select the optimal translation.
\citet{chen2024cratmultiagentframeworkcausalityenhanced} highlight persistent challenges in LLM-based translation with domain-specific terminology, and propose CRAT, a framework combining Retrieval-Augmented Generation (RAG) and causality-enhanced reflection. By coordinating specialized agents for unknown term detection, knowledge graph construction, causal validation, and translation generation, CRAT significantly improves translation consistency in complex and evolving contexts.
Recent work such as \citet{wu-etal-2024-transagents} and \citet{wu2024perhapshumantranslationharnessing} explores multi-agent frameworks that emulate the traditional translation publication process. However, \citet{wu2024perhapshumantranslationharnessing} deploys over 30 specialized agents, introducing considerable system complexity and making it difficult to discern which agents substantively contribute to translation quality. Moreover, their evaluation largely depends on win-rate comparisons based on the preferences of humans and LLMs, lacking objective, standardized metrics necessary for a more rigorous and reproducible assessment.

In the domain of translation quality evaluation, \citet{feng2025mmadmultidimensionalmultiagentdebate} decomposes the MQM~\citep{mqm_error_typology} error typology into four dimensions---Accuracy, Fluency, Style, and Terminology---followed by multi-agent debates and a final judgment phase. While sharing some conceptual similarity with our work, their framework adheres closely to the original MQM structure, resulting in a relatively complex and fragmented evaluation process. In contrast, we reframe MQM into three linguistically grounded dimensions---\textit{faithfulness}, \textit{expressiveness}, and \textit{elegance}---achieving a better balance between interpretability and operational simplicity. Additionally, the inclusion of debate and judgment stages introduces considerable evaluation overhead compared to our more streamlined and efficient design.

\section{Conclusion, Limitations, and Future Work}
We propose TACTIC, a multi-agent translation framework inspired by Cognitive Translation Studies (CTS), which simulates six distinct cognitive roles: drafting, refinement, evaluation, scoring, context reasoning, and additional knowledge gathering. TACTIC supports both a base workflow and a complex workflow to accommodate varying translation scenarios. Experiments across multiple model families and evaluation metrics show that TACTIC significantly improves translation quality, achieving state-of-the-art performance on the FLORES-200 and WMT24 test sets.

While our framework achieves notable improvements in translation quality, it currently relies solely on automatic evaluation metrics, which may not fully align with human judgments. Additionally, due to its multi-stage nature, the multi-agent workflow incurs higher latency compared to direct translation. Reducing inference time while preserving accuracy remains a key direction for future research. Furthermore, our current integration of cognitive translation theory is still preliminary. We plan to deepen our exploration of Cognitive Translation Studies and incorporate a broader range of theoretical foundations into the agent design of translation systems in future work.

\section{Acknowledgments}
We would like to express our sincere gratitude to Tianle Zhou of Wuhan Yangtze Computing Technology Co., Ltd. for his support in deploying the Ascend 910B cluster.

\medskip
\bibliographystyle{plainnat}
\bibliography{neurips_2025}
\newpage

\appendix

\section{Additional Results}
\label{Additional Results}

As shown in Table~\ref{tab:main results chrf bleu metricx}, Lexical-based metrics such as BLEU and ChrF exhibit clear inconsistencies compared to model-based metrics like XCOMET, COMETKIWI-23, and MetricX-24. Consistent with the findings of~\citet{deutsch2025wmt24++}, model-based metrics demonstrate strong internal agreement, consistently assigning higher scores to leading LLMs (e.g., OpenAI o1, Claude, and Gemini), and aligning more closely with human evaluations. In contrast, lexical-based metrics often favor traditional machine translation systems such as \textit{Google Translate} and \textit{DeepL Translate}. Therefore, we include BLEU and ChrF results primarily for completeness.

\input{Analysis/main_chrf_bleu_metricx}
\input{Analysis/case_study_2}
\input{Analysis/en_zh_comet_metricx}

\begin{table}[htbp]
  \centering
  \scriptsize
  \caption{Comparison results of the TACTIC framework under zero-shot and few-shot (default) prompt settings, applied to the DraftAgent and RefinementAgent using the Qwen3-32B model in non-inference mode. Overall, few-shot performs better for en-zh, while zero-shot is superior for zh-en.}
  \label{tab:qwen3_32b_zeroshot_fewshot_comparison}
  \begin{tabular}{lllcccccc}
  \toprule
    \textbf{Prompts} & \textbf{Directions} & \textbf{Datasets} & \textbf{ChrF} & \textbf{BLEU} & \textbf{TER} & \textbf{XCOMET} & \textbf{COMETKIWI-23} & \textbf{MetricX-24} \\
  \midrule
  \multirow{4}{*}{zero-shot} 
    & \multirow{2}{*}{en-zh} & FLORES-200 & 39.01 & 45.84 & 100.14 & 93.78 & 89.18 & -1.74 \\
    &                        & WMT24     & 37.05 & 38.81 & 102.09 & 85.51 & 78.47 & -2.55 \\
    & \multirow{2}{*}{zh-en} & FLORES-200 & 59.41 & 28.13 & 61.59  & 96.97 & 88.89 & -0.97 \\
    &                        & WMT24     & 53.28 & 23.64 & 67.32  & 89.29 & 79.50 & -2.26 \\
  \midrule
  \multirow{4}{*}{few-shot} 
    & \multirow{2}{*}{en-zh} & FLORES-200 & 38.52 & 45.30 & 99.95  & 94.16 & 89.60 & -1.70 \\
    &                        & WMT24     & 36.70 & 38.47 & 98.75  & 85.87 & 79.46 & -2.45 \\
    & \multirow{2}{*}{zh-en} & FLORES-200 & 59.14 & 27.65 & 61.92  & 96.86 & 88.79 & -0.99 \\
    &                        & WMT24     & 53.25 & 23.40 & 67.79  & 88.55 & 79.18 & -2.38 \\
  \bottomrule
  \end{tabular}
\end{table}


\section{Implementation Details}
\label{Implementation Details}
\vspace{-0.8em}
All open-source models, except DeepSeek-V3 and DeepSeek-R1, are primarily deployed locally using \texttt{vLLM}~\citep{kwon2023efficient} on machines each equipped with 8~\(\times\)~NVIDIA A100 GPUs (40GB per GPU). In contrast, DeepSeek-V3 and DeepSeek-R1 are deployed in a distributed manner across multiple machines, each equipped with 8~\(\times\)~Ascend 910B NPUs (64GB per NPU). For all models except DeepSeek-V3, the output parameters are consistently set to \texttt{max\_model\_len} = 8192, \texttt{max\_tokens} = 4096, and \texttt{temperature} = 0.6. For DeepSeek-V3, we follow the official recommendation and adopt a lower temperature of 0.3. Due to the inherent randomness of LLMs outputs, individual experimental results may exhibit slight fluctuations, however, the overall distribution remains stable.

In edge-case scenarios or when processing syntactically complex inputs, TACTIC may enter a prolonged iterative cycle. To ensure the framework remains robust across diverse conditions and produces translations within a controllable time frame, we introduce two additional constraints: a maximum iteration threshold ($\kappa$) and a maximum execution time threshold ($\delta$). Once either threshold is reached, the framework outputs the highest-scoring translation generated during the iterative process.

\section{Error Analysis}
\label{Error Analysis}
We encountered several practical issues while experimenting with our framework. Below is a concise summary:

\begin{itemize}[leftmargin=*]
\item \textbf{External API instability}: When using external APIs as backend models, we occasionally experienced request failures. To improve robustness, we implemented an automatic retry mechanism upon failure.

\item \textbf{Strict JSON formatting requirements}: Given that our agent framework involves extensive data parsing, strict control over JSON-formatted outputs is necessary. For external APIs, we manually verify whether JSON output is supported via their official documentation. For local deployment, we utilize structured output support provided by \textit{vLLM}.

\item \textbf{Repetitive output in local inference}: When running inference with locally deployed models, we occasionally observed endless repetition in output. For example, on a WMT24 test sample "@user47 noooooOOOOooOoOooooo", smaller models (e.g., 7B, 14B) may repeatedly output "@user47 noooooOOOOooOoOooo..." until the maximum token length is reached. To mitigate this, we implemented an overlength detection and retry mechanism.

\item \textbf{Untranslated inputs}: For certain source texts such as "@user8", smaller models may fail to generate valid translations. In such cases, we enable a retry mechanism to ensure output quality.
\end{itemize}


\section{Prompts}
\label{Prompts}

\raisebox{-.4\baselineskip}{\includegraphics[height=1.3\baselineskip]{figs/agents/drafting-agent.png}}~~\textbf{\textit{DraftAgent}}
\begin{lstlisting}
SYSTEM_PROMPT
You are a native speaker of both {source_language} and {target_language}, with expertise in translating from {source_language} to {target_language}.

USER_PROMPT
In this phase, your sole objective is to generate a draft translation that strictly adheres to the source text. Avoid adding any additional information not present in the source text, nor omit any content from it. Every detail must be fully preserved in the {target_language} translation. Below are several translation strategies. Please provide your best {translation_type} for the following source text.

Translation Strategies:
1. Literal Translation: Also known as direct translation or word-for-word translation, it prioritizes accurate meaning while preserving the original text's form and content in the target language.  
2.Sense-for-Sense Translation: Focuses on conveying the core meaning of the original text without adhering strictly to its linguistic form, ensuring greater fluency and naturalness in the target language.  
3.Free Translation: Emphasizes delivering the overall meaning and effect of the original text, allowing for significant rewriting or restructuring as needed.

## Pre-translation Research:
{pre_translation_result}

## Context Analysis:
{context_analysis}

## Extended Context:
{extended_context}

## Few-shot Examples:
{few_shot_examples}

## Source Text ({source_language}):
{source_text}

## Output format specification:
```json
{{
    "translation": "<Your translation>"
}}
```
The JSON object: json
\end{lstlisting}

\raisebox{-.4\baselineskip}{\includegraphics[height=1.3\baselineskip]{figs/agents/refining-agent.png}}~~\textbf{\textit{RefinementAgent}}
\begin{lstlisting}
SYSTEM_PROMPT
You are a native speaker of both {source_language} and {target_language}, with expertise in translating from {source_language} to {target_language}. As an assistant dedicated to enhancing translation quality, you will be given a source sentence in {source_language}, a list of candidate translations in {target_language}, along with relevant research. Your task is to carefully analyze the provided information and refine the translation, ensuring it accurately and fully captures the original meaning of the source text. Your analysis should be in English.

USER_PROMPT
Refine the translation from {source_language} to {target_language}.

## Pre-translation Research:
{pre_translation_result}

## Context Analysis:
{context_analysis}

## Extended Context:
{extended_context}

## Few-shot Examples:
{few_shot_examples}

## Source text ({source_language}):
{source_text}

## Candidate translations ({target_language}): 
{candidate_translations}

## Output format specification:
```json
{{
    "analysis": "<Brief analysis of the candidate translations>",
    "translation": "<Your refined translation>"
}}
```
The JSON object: json
\end{lstlisting}

\raisebox{-.4\baselineskip}{\includegraphics[height=1.3\baselineskip]{figs/agents/evaluating-agent.png}}~~\textbf{\textit{EvaluationAgent}}
\begin{lstlisting}
SYSTEM_PROMPT
You are a native speaker of both {source_language} and {target_language}, as well as an expert in translation. Given a source sentence and its translation, your task is to assess the quality of the translation and provide suggestions for improvement.

USER_PROMPT
Please evaluate the faithfulness, expressiveness and elegance of the given translation based on the provided criteria:

Faithfulness Evaluation Criteria: 
1.Addition: Translation includes information not present in the source. 
2.Omission: Translation is missing content from the source. 
3.Mistranslation: Translation does not accurately represent the source. 
4.Untranslated text: Source text has been left untranslated. 

Expressiveness Evaluation Criteria:
1.Punctuation: Incorrect punctuation (for locale or style).
2.Spelling: Incorrect spelling or capitalization.
3.Grammar: Problems with grammar, other than orthography.
4.Register: Wrong grammatical register (eg, inappropriately informal pronouns).
5.Inconsistency: Internal inconsistency (not related to terminology).
6.Character encoding: Characters are garbled due to incorrect encoding.

Elegance Evaluation Criteria:
1.Terminology: Terminology is either non-standard, does not fit the context, or is used inconsistently.
2.Style: Translation has stylistic problems.
3.Locale convention: Wrong format for addresses, currency, dates, names, telephone numbers, time expressions, or other locale-specific elements.
4.Logical Expression: Translation lacks logical coherence or does not align with the thinking patterns and language expressions of the target language.
5.Other: Any other issues that might affect the elegance of the translation.

## Source Text ({source_language}):
{source_text}

## Translation ({target_language}):
{translation}

## Output format specification:
```json
{{
    "faithfulness": "<Your assessment based on the faithfulness criteria.>",
    "expressiveness": "<Your assessment based on the expressiveness criteria.>",
    "elegance": "<Your assessment based on the elegance criteria.>"
}}
```
The JSON object: json
\end{lstlisting}

\raisebox{-.4\baselineskip}{\includegraphics[height=1.3\baselineskip]{figs/agents/scoring-agent.png}}~~\textbf{\textit{ScoreAgent}}
\begin{lstlisting}
SYSTEM_PROMPT
You are a native speaker of both {source_language} and {target_language}, as well as an expert in translation. As an assistant dedicated to enhancing translation quality, you will be provided with a source sentence in {source_language}, its translation in {target_language}, and three evaluation criteria. Your task is to assess the translation based on three factors: faithfulness (10 points), expressiveness (10 points), and elegance (10 points). Each score ranges from 1 to 10, where lower scores indicate poorer quality, and higher scores indicate better quality. A score of 10 indicates that the translation is flawless in that specific dimension.

USER_PROMPT
Please score the translation based on the following criteria, evaluations, and source text.

Faithfulness Evaluation Criteria:
1. Addition: Translation includes information not present in the source.
2. Omission: Translation is missing content from the source.
3. Mistranslation: Translation does not accurately represent the source.
4. Untranslated text: Source text has been left untranslated.

Expressiveness Evaluation Criteria:
1. Punctuation: Incorrect punctuation (for locale or style).
2. Spelling: Incorrect spelling or capitalization.
3. Grammar: Problems with grammar, other than orthography.
4. Register: Wrong grammatical register (e.g., inappropriately informal pronouns).
5. Inconsistency: Internal inconsistency (not related to terminology).
6. Character encoding: Characters are garbled due to incorrect encoding.

Elegance Evaluation Criteria:
1. Terminology: Terminology is either non-standard, does not fit the context, or is used inconsistently.
2. Style: Translation has stylistic problems.
3. Locale convention: Wrong format for addresses, currency, dates, names, telephone numbers, time expressions, or other locale-specific elements.
4. Logical Expression: Translation lacks logical coherence or does not align with the thinking patterns and language expressions of the target language.
5. Other: Any other issues that might affect the elegance of the translation.

## Source Text ({source_language}):
{source_text}

## Translation ({target_language}):
{translation}

## Evaluation:
{evaluation_result}

## Output format specification:
```json
{{
    "faithfulness_score": "<faithfulness score>",
    "expressiveness_score": "<expressiveness score>",
    "elegance_score": "<elegance score>",
    "overall_score": "<The sum of faithfulness_score, expressiveness_score and elegance_score.>",
    "feedback": "<Brief feedback for each criterion, explaining your score.>"
}}
```
The JSON object: json
\end{lstlisting}


\raisebox{-.4\baselineskip}{\includegraphics[height=1.3\baselineskip]{figs/agents/context-agent.png}}~~\textbf{\textit{ContextAgent}}
\begin{lstlisting}
SYSTEM_PROMPT
You are a native speaker of both {source_language} and {target_language}, with expertise in translating from {source_language} to {target_language}.

USER_PROMPT
Below is a source sentence in {source_language}. Your task is to infer the possible context, including style, purpose, target audience and other environmental factors that maybe helpful in understanding the source text, Your context analysis should be in English. Then, expand the source sentence by adding a previous sentence and a next sentence, using the same language, together, these sentences should form a coherent and complete passage. 

## Source Text ({source_language}):
{source_text}

## Output format specification:
```json
{{
    "context_analysis": "<Your context analysis>",
    "extended_context": "Expanded previous sentence. source sentence. Expanded next sentence."
}}
```
The JSON object: json
\end{lstlisting}

\raisebox{-.4\baselineskip}{\includegraphics[height=1.3\baselineskip]{figs/agents/research-agent.png}}~~\textbf{\textit{ResearchAgent}}
\begin{lstlisting}
SYSTEM_PROMPT
You are a native speaker of both {source_language} and {target_language}, with expertise in translating from {source_language} to {target_language}.

USER_PROMPT
Before translation, conducting thorough pre-translation research is crucial to identifying elements of the text that may pose translation challenges. The objective at this stage is to compile a list of keywords and phrases essential for accurately understanding the source text. These may include technical terms, proper names, idiomatic expressions, or other context-dependent elements that require special attention to ensure precise translation.

## Source Text ({source_language}):
{source_text}

Please list keywords and phrases directly in {source_language} and {target_language} using the following output format. Do not generate any other content:  

1. keyword1: keyword1 in {target_language}  
2. phrases2: phrases2 in {target_language}
\end{lstlisting}

\raisebox{-.4\baselineskip}~\textbf{\textit{Zero-shot}}
\begin{lstlisting}
SYSTEM_PROMPT
You are a helpful assistant.
USER_PROMPT
Translate this from {source_language} to {target_language}:
{source_language}: {source_text}
{target_language}:

Please generate the final translation in JSON format as follows:
## Output format specification:
```json
{{
    "translation": "<Your translation>"
}}
```
The JSON object: json
\end{lstlisting}

\newpage

\raisebox{-.4\baselineskip}~\textbf{\textit{Few-shot}}
\begin{lstlisting}
SYSTEM_PROMPT
You are a helpful assistant.
USER_PROMPT
Translate this from {source_language} to {target_language}:
{few_shot_examples}

{source_language}: {source_text}
{target_language}:

Please generate the final translation in JSON format as follows:
## Output format specification:
```json
{{
    "translation": "<Your translation>"
}}
```
The JSON object: json
\end{lstlisting}

%% file: Analysis/main_result.tex
\begin{table}[tbp]
\centering
\tiny
\caption{Evaluation results of XCOMET and COMETKIWI-23 (abbreviated as KIWI-23) scores across different models on the FLORES-200 and WMT24 test sets. \textit{Models} are employed for non-translation tasks (e.g., evaluation and score), whereas \textit{Trans Models} are utilized for translation tasks (draft and refinement). All Qwen2.5 series models are Instruct versions and the ``Instruct'' suffix is omitted in the table for brevity. The best scores are highlighted in bold, and the second-best scores are underlined.}

\label{tab:main results}
\begin{tabularx}{\textwidth}{
>{\centering\arraybackslash}l
>{\centering\arraybackslash}l
*{8}{>{\centering\arraybackslash}X}
} 
\toprule
&  & \multicolumn{4}{c}{\textbf{FLORES-200}} & \multicolumn{4}{c}{\textbf{WMT24}} \\
 \cmidrule(lr){3-6} \cmidrule(lr){7-10}
 {\textbf{Models}} & {\textbf{Trans Models}}& \multicolumn{2}{c}{\textbf{en$\rightarrow$xx}} & \multicolumn{2}{c}{\textbf{xx$\rightarrow$en}}  & \multicolumn{2}{c}{\textbf{en$\rightarrow$xx}} & \multicolumn{2}{c}{\textbf{xx$\rightarrow$en}} \\
\cmidrule(lr){3-4} \cmidrule(lr){5-6} \cmidrule(lr){7-8} \cmidrule(lr){9-10} & & {\textbf{XCOMET}} & {\textbf{KIWI-23}} & {\textbf{XCOMET}} & {\textbf{KIWI-23}} & {\textbf{XCOMET}} & {\textbf{KIWI-23}} & {\textbf{XCOMET}} & {\textbf{KIWI-23}} \\
    \midrule
    \specialrule{0em}{1pt}{1pt}
    \rowcolor{gray!30}
    \multicolumn{10}{c}{\textbf{\textit{Zero-shot}}} \\
    \specialrule{0em}{1pt}{1pt}
    \hline
    \specialrule{0em}{1pt}{1pt}
    \multicolumn{10}{c}{\textbf{\textit{Non-reasoning models}}} \\
    \specialrule{0em}{1pt}{1pt}
    
     Qwen2.5-7B & Qwen2.5-7B & 77.53 & 70.18 & 91.55 & 85.12 & 66.97 & 59.97 & 81.21 & 77.11 \\
     Qwen2.5-14B & Qwen2.5-14B & 84.17 & 78.25 & 94.05 & 87.17 & 73.97 & 68.57 & 82.59 & 77.83 \\
     Qwen2.5-32B & Qwen2.5-32B & 86.97 & 81.27 & 94.43 & 87.43 & 76.21 & 70.87 & 84.37 & 79.40 \\
     Qwen2.5-72B & Qwen2.5-72B & 91.45 & 86.14 & 95.71 & 88.20 & 81.23 & 75.81 & 86.00 & 80.48 \\
     GPT-4.1 & GPT-4.1 & 95.46 & 90.83 & \underline{96.50} & 88.61 & 86.83 & 81.03 & 87.73 & 81.33 \\
     DeepSeek-V3 & DeepSeek-V3 & 94.43 & 89.80 & 96.14 & 88.45 & 84.71 & 79.61 & 87.20 & 80.99 \\
    \specialrule{0em}{1pt}{1pt}
    \hdashline
    \specialrule{0em}{1pt}{1pt}
    \multicolumn{10}{c}{\textbf{\textit{Reasoning-capable models}}} \\
    \specialrule{0em}{1pt}{1pt}
     QwQ-32B & QwQ-32B & 90.87 & 86.33 & 95.02 & 87.82 & 79.92 & 75.42 & 83.59 & 78.74 \\
     DeepSeek-R1 & DeepSeek-R1 & 95.33 & 90.75 & 96.13 & 88.36 & 86.10 & 80.45 & \underline{87.98} & 75.01 \\
    \hline
    \specialrule{0em}{1pt}{1pt}
    \rowcolor{gray!30}
    \multicolumn{10}{c}{\textbf{\textit{Few-shot}}} \\
    \specialrule{0em}{1pt}{1pt}
    \hline
    \specialrule{0em}{1pt}{1pt}
     Qwen2.5-7B & Qwen2.5-7B & 77.50 & 69.73 & 91.06 & 85.00 & 65.25 & 59.03 & 78.44 & 75.09 \\
     Qwen2.5-14B & Qwen2.5-14B & 85.44 & 79.63 & 94.09 & 87.21 & 75.03 & 69.62 & 82.66 & 77.93 \\
     Qwen2.5-32B & Qwen2.5-32B & 87.65 & 81.87 & 94.31 & 87.28 & 76.61 & 71.22 & 84.36 & 79.48 \\
     Qwen2.5-72B & Qwen2.5-72B & 91.80 & 86.34 & 95.70 & 88.22 & 81.32 & 76.04 & 85.89 & 80.42 \\
     GPT-4.1 & GPT-4.1 & 95.59 & 91.06 & 96.45 & 88.60 & 86.92 & 81.07 & 87.90 & 81.39 \\
     DeepSeek-V3 & DeepSeek-V3 & 94.90 & 90.25 & 96.24 & 88.50 & 85.58 & 80.09 & 87.05 & 81.04 \\
    \hline
    \specialrule{0em}{1pt}{1pt}
    \rowcolor{gray!30}
    \multicolumn{10}{c}{\textbf{\textit{TACTIC}}} \\
    \specialrule{0em}{1pt}{1pt}
    \hline
    \specialrule{0em}{1pt}{1pt}
    \multicolumn{10}{c}{\textbf{\textit{Non-reasoning models}}} \\
    \specialrule{0em}{1pt}{1pt}
     Qwen2.5-7B & Qwen2.5-7B & 87.69 & 82.01 & 94.01 & 88.44 & 73.13 & 68.15 & 83.66 & 78.07 \\
     Qwen2.5-14B & Qwen2.5-14B & 91.33 & 86.68 & 95.22 & 89.39 & 81.12 & 76.25 & 86.10 & 80.74 \\
     Qwen2.5-32B & Qwen2.5-32B & 93.20 & 89.17 & 95.86 & \underline{89.80} & 82.97 & 78.06 & 87.13 & \underline{81.56} \\
     Qwen2.5-72B & Qwen2.5-72B & 93.48 & 88.72 & 95.82 & 88.26 & 82.71 & 78.08 & 86.17 & 80.35 \\
     DeepSeek-V3 & DeepSeek-V3 & \textbf{96.19} & \textbf{92.64} & \textbf{96.69} & \textbf{90.15} & 86.95 & \underline{81.26} & \textbf{89.07} & \textbf{82.46} \\
    \specialrule{0em}{1pt}{1pt}
    \hdashline
    \specialrule{0em}{1pt}{1pt}
    \multicolumn{10}{c}{\textbf{\textit{Reasoning-capable models}}} \\
    \specialrule{0em}{1pt}{1pt}
     Qwen2.5-7B & QwQ-32B & 92.56 & 88.22 & 95.37 & 88.00 & 81.68 & 77.38 & 85.85 & 80.20 \\
     Qwen2.5-14B & QwQ-32B & 92.90 & 88.51 & 95.70 & 88.19 & 83.47 & 78.65 & 87.20 & 81.54 \\
     Qwen2.5-32B & QwQ-32B & 93.11 & 88.69 & 95.65 & 88.17 & 82.69 & 77.83 & 86.34 & 80.54 \\
     Qwen2.5-72B & QwQ-32B & 93.38 & 88.97 & 95.72 & 88.18 & 82.95 & 78.34 & 86.36 & 80.52 \\
     Qwen2.5-32B & DeepSeek-R1 & \underline{95.95} & \underline{91.72} & 96.10 & 88.01 & \textbf{87.60} & \textbf{81.78} & 87.88 & 81.19 \\
     DeepSeek-V3 & DeepSeek-R1 & 95.78 & 91.32 & 96.06 & 88.08 & \underline{87.10} & 81.08 & 87.62 & 80.81 \\
    \bottomrule
  \end{tabularx}
\end{table}

%% file: Analysis/case_study_1.tex
\begin{figure}[tbp]
\centering
\includegraphics[width=\linewidth]{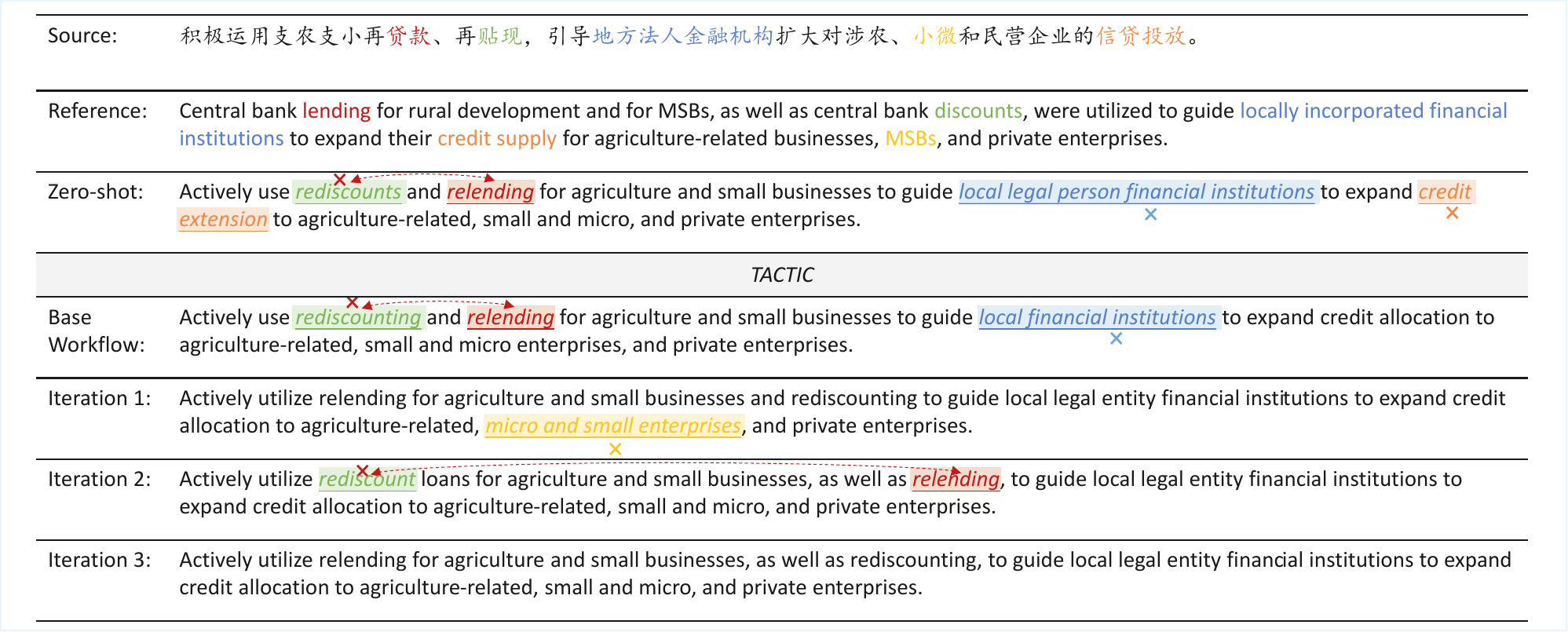}
\caption{
Case study demonstrating translation refinement through iterative agent collaboration. The backend model is Qwen2.5-32B-Instruct.
}
\label{fig: case_study_1}
\end{figure}

%% file: Analysis/ana1.tex
\paragraph{Synergistic Effects of TACTIC and Reasoning-Capable Models.} We compare two experimental settings: (1) a TACTIC-based configuration employing DeepSeek-V3 and DeepSeek-R1; and (2) a zero-shot baseline using DeepSeek-R1 for both roles. TACTIC achieves higher scores on both XCOMET (95.78 vs. 95.33) and COMETKIWI-23 (91.32 vs. 90.75), demonstrating clear advantages. This improvement stems from TACTIC’s structured, agent-based workflow, which enhances the deployment of reasoning capabilities by assigning focused sub-tasks to each agent. The collaborative process encourages deeper understanding of source content, enabling better fidelity, fluency, and elegance in translation.

\vspace{-0.8em}
\paragraph{Adaptability of Reasoning-Capable Models Within TACTIC.} To isolate the impact of reasoning ability and framework structure, we compare four settings varying in model type (DeepSeek-R1 vs. DeepSeek-V3) and translation approach (zero-shot vs. TACTIC). In zero-shot settings, the reasoning-capable R1 outperforms V3. However, under TACTIC, V3 shows the greatest performance gain (XCOMET: 96.19), surpassing even the reasoning-enhanced TACTIC setup. This suggests that the benefits of structured collaboration outweigh intrinsic reasoning capabilities, especially for weaker models. TACTIC thus acts as a compensatory mechanism, amplifying translation quality by simulating human-like decision-making processes.

\vspace{-0.8em}
\paragraph{Effectiveness of Iterative Refinement in TACTIC: A Case Study.} We examine a case from a central bank monetary policy report in Figure~\ref{fig: case_study_1}, characterized by dense terminology and syntactic ambiguity. The baseline system struggles due to domain complexity and structural divergence between source and reference. In contrast, TACTIC leverages modular agents to identify and resolve terminology, context, and style mismatches through structured refinement. Each iteration incrementally improves fidelity and fluency, ultimately producing a polished translation aligned with domain expectations. This validates the core design of TACTIC as an effective mechanism for complex, formal-domain translation.

\vspace{-0.8em}
\paragraph{Distributional Impact of Iterative Refinement}
To assess refinement effects, we visualize the score distribution before and after TACTIC's iterative process as shown in Figure~\ref{fig: score_distribution}. Most samples show noticeable gains, while a small subset exhibits slight regression, typically due to over-correction. The overall shift toward higher scores confirms the robustness of the agent-guided refinement mechanism, consistently improving alignment with source intent and target language norms.

\input{Analysis/xcomet_score_distribution}


\vspace{-0.8em}
\paragraph{Divergence Across Evaluation Dimensions}
To evaluate sensitivity to nuanced translation challenges, we analyze an idiomatic example in Appendix Figure~\ref{fig: case_study_2}: "That idea went out the window." Literal, interpretive, and free strategies produce divergent outputs. RefinementAgent correctly selects the most idiomatic and contextually appropriate version. This case highlights that strategy diversity provides valuable inputs for downstream refinement, especially in figurative or culturally loaded contexts. EvaluationAgent assigns perfect scores (10/10/10) across faithfulness, expressiveness, and elegance—demonstrating its ability to discern subtle quality variations across multiple dimensions.

%% file: Analysis/xcomet_score_distribution.tex
\begin{figure}[tbp]
\centering
\begin{minipage}{0.33\textwidth}
\includegraphics[width=\linewidth]{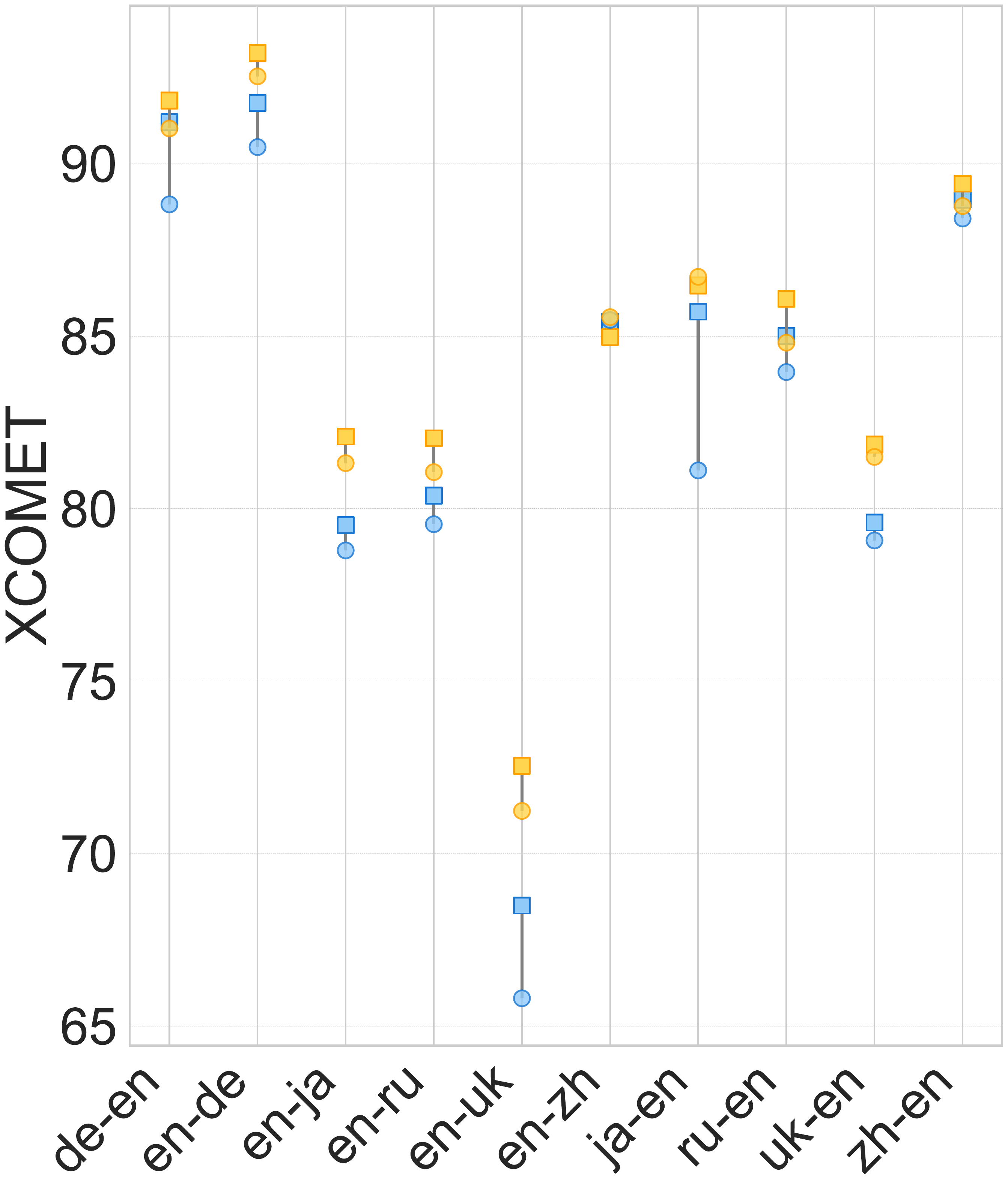}
\end{minipage}
\begin{minipage}{0.32\textwidth}
\includegraphics[width=\linewidth]{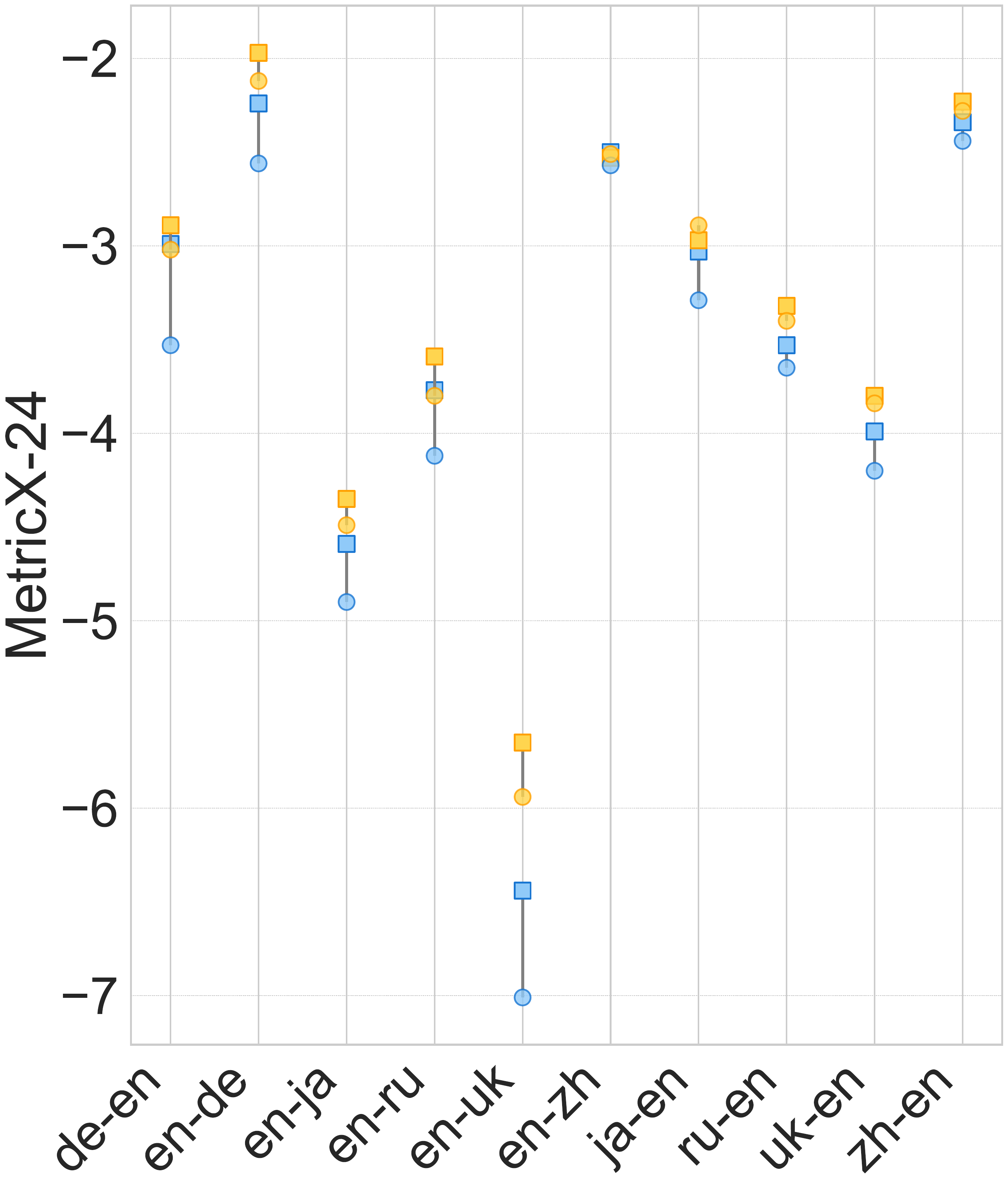}
\end{minipage}
\begin{minipage}{0.33\textwidth}
\includegraphics[width=\linewidth]{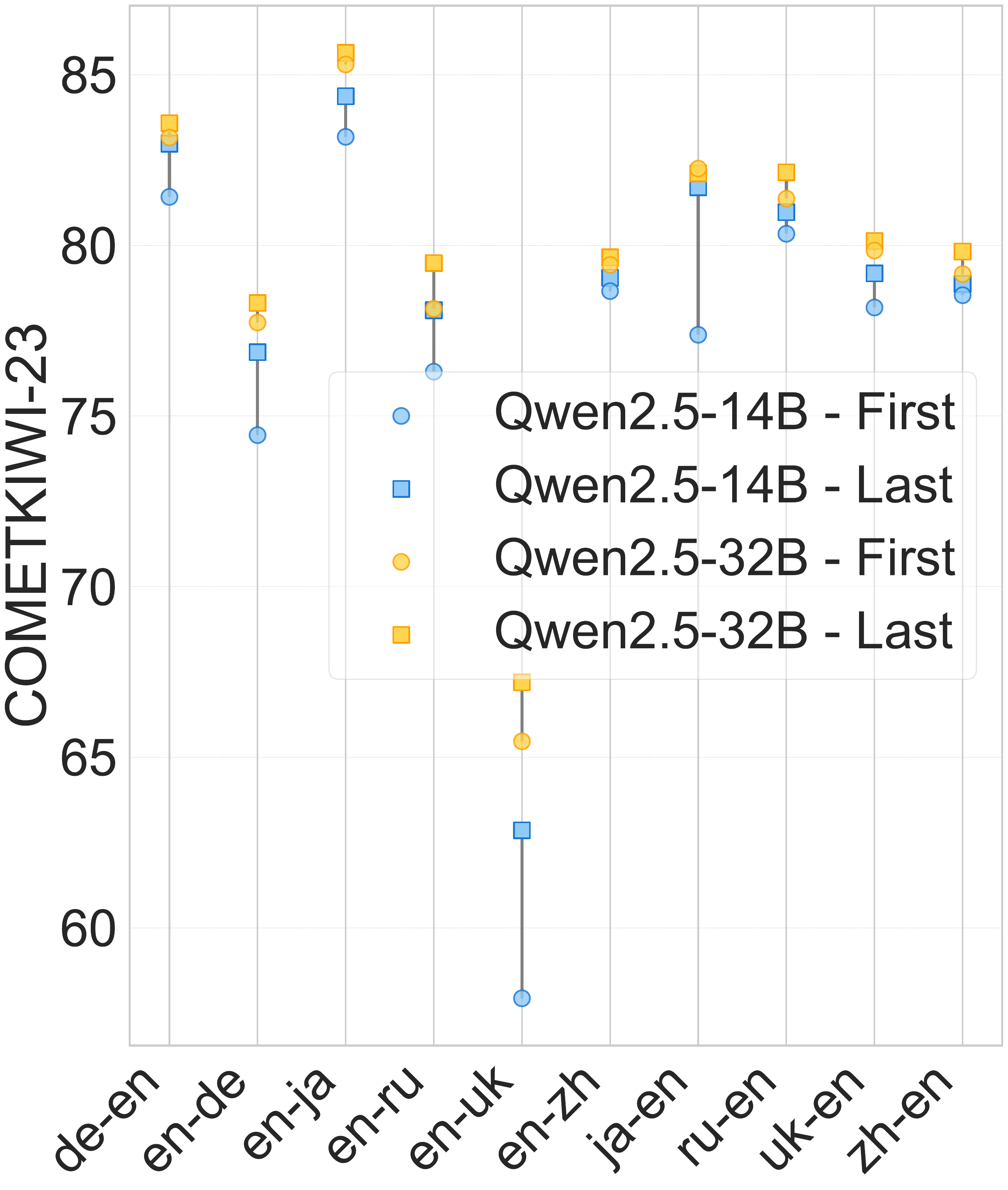}
\end{minipage}
\caption{
Distributional Impact of Iterative Refinement on the WMT24 Test Set.
``Fist'' and ``'Last'' denote the performance at the initial and final epochs, respectively. 
}
\vspace{-0.3cm}
\label{fig: score_distribution}
\end{figure}

%% file: Analysis/main_chrf_bleu_metricx.tex
\begin{table}[htbp]
\centering
\tiny
\caption{Evaluation results of ChrF, BLEU, and MetricX-24 (abbreviated as MetricX) scores across different models on the FLORES-200 and WMT24 test sets. \textit{Models} are employed for non-translation tasks (e.g., evaluation and score), whereas \textit{Trans Models} are utilized for translation tasks (draft and refinement). All Qwen2.5 series models are Instruct versions and the ``Instruct'' suffix is omitted in the table for brevity. The best scores are highlighted in bold, and the second-best scores are underlined.}

\label{tab:main results chrf bleu metricx}
\begin{tabularx}{\textwidth}{
>{\centering\arraybackslash}l
>{\centering\arraybackslash}l
*{12}{>{\centering\arraybackslash}X}
} 
\toprule
&  & \multicolumn{6}{c}{\textbf{FLORES-200}} & \multicolumn{6}{c}{\textbf{WMT24}} \\
 \cmidrule(lr){3-8} \cmidrule(lr){9-14}
 {\textbf{Models}} & {\textbf{Trans Models}}& \multicolumn{3}{c}{\textbf{en$\rightarrow$xx}} & \multicolumn{3}{c}{\textbf{xx$\rightarrow$en}}  & \multicolumn{3}{c}{\textbf{en$\rightarrow$xx}} & \multicolumn{3}{c}{\textbf{xx$\rightarrow$en}} \\
\cmidrule(lr){3-5} \cmidrule(lr){6-8} \cmidrule(lr){9-11} \cmidrule(lr){12-14}  & & {\textbf{ChrF}} & {\textbf{BLEU}} & {\textbf{MetricX}} & {\textbf{ChrF}} & {\textbf{BLEU}} & {\textbf{MetricX}} & {\textbf{ChrF}} & {\textbf{BLEU}} & {\textbf{MetricX}} & {\textbf{ChrF}} & {\textbf{BLEU}}& {\textbf{MetricX}} \\
    \midrule
    \specialrule{0em}{1pt}{1pt}
    \rowcolor{gray!30}
    \multicolumn{14}{c}{\textbf{\textit{Zero-shot}}} \\
    \specialrule{0em}{1pt}{1pt}
    \hline
    \specialrule{0em}{1pt}{1pt}
    \multicolumn{14}{c}{\textbf{\textit{Non-reasoning models}}} \\
    \specialrule{0em}{1pt}{1pt}
    
     Qwen2.5-7B & Qwen2.5-7B & 42.55 & 25.32 & -5.66 & 59.60 & 31.25 & -2.64 & 38.59 & 21.44 & -6.91 & 52.35 & 24.41 & -4.04 \\
     Qwen2.5-14B & Qwen2.5-14B & 46.24 & 28.85 & -4.07 & 62.14 & 33.28 & -2.05 & 42.35 & 24.35 & -5.11 & 54.12 & 26.73 & -3.76 \\
Qwen2.5-32B & Qwen2.5-32B & 48.44 & 31.28 & -3.49 & 62.68 & 34.12 & -2.00 & 43.43 & 25.58 & -4.75 & 54.87 & 27.15 & -3.36 \\
Qwen2.5-72B & Qwen2.5-72B & 51.04 & 33.89 & -2.64 & 64.00 & 36.02 & -1.73 & 46.10 & 28.25 & -3.81 & 56.17 & 28.32 & -3.02 \\
GPT-4.1 & GPT-4.1 & \textbf{55.45} & \underline{39.00} & -1.76 & \textbf{64.99} & 36.35 & \textbf{-1.57} & \underline{49.65} & 31.51 & -2.81 & 57.50 & 29.06 & \underline{-2.73} \\
DeepSeek-V3 & DeepSeek-V3 & 55.22 & 38.84 & -1.94 & 64.42 & \underline{36.46} & -1.68 & \textbf{49.70} & \textbf{32.41} & -3.11 & 57.32 & \underline{29.58} & -2.87 \\
    \specialrule{0em}{1pt}{1pt}
    \hdashline
    \specialrule{0em}{1pt}{1pt}
    \multicolumn{14}{c}{\textbf{\textit{Reasoning-capable models}}} \\
    \specialrule{0em}{1pt}{1pt}
QwQ-32B & QwQ-32B & 46.48 & 26.00 & -2.69 & 61.47 & 30.98 & -1.86 & 40.88 & 19.77 & -3.99 & 54.54 & 26.57 & -3.45 \\
DeepSeek-R1 & DeepSeek-R1 & 53.37 & 35.98 & -1.77 & 63.16 & 34.04 & -1.62 & 46.78 & 28.08 & -2.88 & \textbf{71.59} & \textbf{51.72} & \textbf{-2.13} \\

    \hline
    \specialrule{0em}{1pt}{1pt}
    \rowcolor{gray!30}
    \multicolumn{14}{c}{\textbf{\textit{Few-shot}}} \\
    \specialrule{0em}{1pt}{1pt}
    \hline
    \specialrule{0em}{1pt}{1pt}
    
Qwen2.5-7B & Qwen2.5-7B & 42.35 & 25.20 & -5.83 & 59.50 & 30.63 & -2.73 & 38.16 & 20.56 & -7.14 & 51.49 & 23.56 & -4.40 \\
Qwen2.5-14B & Qwen2.5-14B & 46.54 & 29.22 & -3.76 & 61.94 & 33.22 & -2.03 & 42.28 & 24.16 & -4.90 & 53.93 & 26.23 & -3.69 \\
Qwen2.5-32B & Qwen2.5-32B & 48.34 & 31.53 & -3.40 & 62.39 & 34.14 & -2.01 & 43.40 & 25.80 & -4.68 & 55.12 & 27.02 & -3.27 \\
Qwen2.5-72B & Qwen2.5-72B & 51.13 & 34.15 & -2.59 & 63.86 & 35.71 & -1.71 & 45.97 & 28.11 & -3.76 & 56.14 & 28.16 & -3.01 \\
GPT-4.1 & GPT-4.1 & 55.14 & 38.49 & -1.71 & \underline{64.83} & 36.31 & \textbf{-1.57} & 49.61 & 31.41 & -2.77 & \underline{57.62} & 29.38 & -2.73 \\
DeepSeek-V3 & DeepSeek-V3 & \underline{55.26} & \textbf{39.06} & -1.87 & 64.62 & \textbf{37.05} & -1.64 & 49.56 & \underline{32.29} & -3.01 & 57.21 & 29.52 & -2.85 \\
    \hline
    \specialrule{0em}{1pt}{1pt}
    \rowcolor{gray!30}
    \multicolumn{14}{c}{\textbf{\textit{TACTIC}}} \\
    \specialrule{0em}{1pt}{1pt}
    \hline
    \specialrule{0em}{1pt}{1pt}
    \multicolumn{14}{c}{\textbf{\textit{Non-reasoning models}}} \\
    \specialrule{0em}{1pt}{1pt}
Qwen2.5-7B & Qwen2.5-7B & 46.04 & 28.02 & -3.55 & 60.13 & 30.35 & -2.11 & 40.47 & 22.01 & -5.33 & 52.37 & 22.76 & -3.60 \\
Qwen2.5-14B & Qwen2.5-14B & 48.67 & 30.96 & -2.69 & 61.67 & 32.00 & -1.84 & 43.66 & 24.66 & -3.91 & 54.08 & 24.86 & -3.18 \\
Qwen2.5-32B & Qwen2.5-32B & 50.39 & 33.07 & -2.41 & 62.40 & 33.29 & -1.77 & 44.98 & 26.57 & -3.62 & 54.75 & 25.78 & -3.04 \\
Qwen2.5-72B & Qwen2.5-72B & 51.61 & 34.34 & -2.21 & 63.08 & 34.27 & -1.71 & 46.22 & 27.83 & -3.41 & 55.42 & 26.92 & -2.98 \\
DeepSeek-V3 & DeepSeek-V3 & 54.29 & 37.82 & -1.71 & 63.72 & 35.62 & -1.60 & 48.79 & 31.35 & -2.77 & 56.78 & 28.82 & -2.75 \\

    \specialrule{0em}{1pt}{1pt}
    \hdashline
    \specialrule{0em}{1pt}{1pt}
    \multicolumn{14}{c}{\textbf{\textit{Reasoning-capable models}}} \\
    \specialrule{0em}{1pt}{1pt}
Qwen2.5-7B & QwQ-32B & 49.70 & 31.73 & -2.32 & 61.28 & 31.27 & -1.78 & 43.54 & 24.08 & -3.57 & 53.46 & 23.92 & -3.05 \\
Qwen2.5-14B & QwQ-32B & 49.97 & 31.89 & -2.28 & 61.70 & 31.98 & -1.72 & 43.88 & 24.51 & -3.47 & 53.90 & 24.55 & -2.98 \\
Qwen2.5-32B & QwQ-32B & 49.94 & 31.78 & -2.24 & 61.69 & 31.79 & -1.73 & 43.94 & 24.62 & -3.48 & 53.80 & 24.41 & -2.98 \\
Qwen2.5-72B & QwQ-32B & 50.23 & 32.11 & -2.18 & 61.87 & 32.18 & -1.70 & 44.12 & 24.71 & -3.38 & 53.93 & 24.51 & -2.95 \\
Qwen2.5-32B & DeepSeek-R1 & 52.19 & 33.65 & \textbf{-1.62} & 61.75 & 31.96 & -1.61 & 45.10 & 25.03 & \underline{-2.69} & 53.78 & 24.40 & -2.77 \\
DeepSeek-V3 & DeepSeek-R1 & 52.46 & 33.93 & \underline{-1.64} & 61.91 & 32.22 & -1.60 & 45.56 & 25.80 & \textbf{-2.68} & 54.39 & 25.14 & -2.76 \\
\bottomrule
  \end{tabularx}
\end{table}

%% file: Analysis/case_study_2.tex
\begin{figure}[tbp]
\centering
\includegraphics[width=\linewidth]{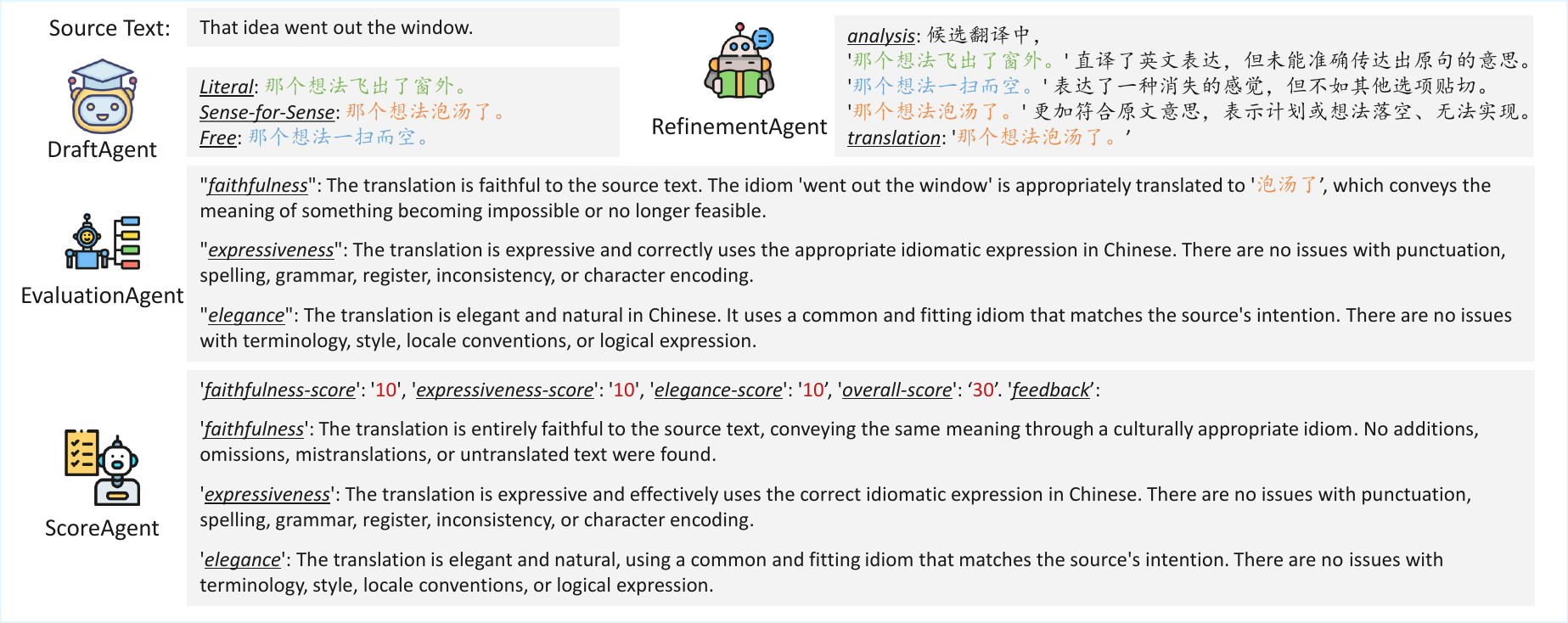}
\caption{
Case study visualizing different translation strategies and evaluation dimensions. The backend model is Qwen2.5-32B-Instruct.
}
\label{fig: case_study_2}
\end{figure}

%% file: Analysis/en_zh_comet_metricx.tex
\begin{table}[htbp]
\centering
\tiny
\caption{Evaluation results (XCOMET and MetricX-24, abbreviated as MetricX) for EN-ZH and ZH-EN translation across different models on the FLORES-200 and WMT24 test sets. \textit{Models} are employed for non-translation tasks (e.g., evaluation and score), whereas \textit{Trans Models} are utilized for translation tasks (draft and refinement). All Qwen2.5 series models are Instruct versions and the ``Instruct'' suffix is omitted in the table for brevity. The best scores are highlighted in bold, and the second-best scores are underlined.}

\label{tab:en-zh results}
\begin{tabularx}{\textwidth}{
>{\centering\arraybackslash}l
>{\centering\arraybackslash}l
*{8}{>{\centering\arraybackslash}X}
} 
\toprule
&  & \multicolumn{4}{c}{\textbf{FLORES-200}} & \multicolumn{4}{c}{\textbf{WMT24}} \\
 \cmidrule(lr){3-6} \cmidrule(lr){7-10}
 {\textbf{Models}} & {\textbf{Trans Models}}& \multicolumn{2}{c}{\textbf{en$\rightarrow$zh}} & \multicolumn{2}{c}{\textbf{zh$\rightarrow$en}}  & \multicolumn{2}{c}{\textbf{en$\rightarrow$zh}} & \multicolumn{2}{c}{\textbf{zh$\rightarrow$en}} \\
\cmidrule(lr){3-4} \cmidrule(lr){5-6} \cmidrule(lr){7-8} \cmidrule(lr){9-10} & & {\textbf{XCOMET}} & {\textbf{MetricX}} & {\textbf{XCOMET}} & {\textbf{MetricX}} & {\textbf{XCOMET}} & {\textbf{MetricX}} & {\textbf{XCOMET}} & {\textbf{MetricX}} \\
    \midrule
    \specialrule{0em}{1pt}{1pt}
    \rowcolor{gray!30}
    \multicolumn{10}{c}{\textbf{\textit{Zero-shot}}} \\
    \specialrule{0em}{1pt}{1pt}
    \hline
    \specialrule{0em}{1pt}{1pt}
    \multicolumn{10}{c}{\textbf{\textit{Non-reasoning models}}} \\
    \specialrule{0em}{1pt}{1pt}
    
    Qwen2.5-7B & Qwen2.5-7B & 88.26 & -2.76 & 92.89 & -1.76 & 79.06 & -3.53 & 85.39 & -2.92 \\
    Qwen2.5-14B & Qwen2.5-14B & 91.52 & -2.04 & 95.22 & -1.30 & 82.07 & -2.83 & 86.03 & -2.63 \\
    Qwen2.5-32B & Qwen2.5-32B & 90.84 & -2.12 & 95.42 & -1.25 & 81.19 & -3.00 & 86.21 & -2.69 \\
    Qwen2.5-72B & Qwen2.5-72B & 92.63 & -1.84 & 96.88 & -0.92 & 83.67 & -2.61 & 87.92 & -2.22 \\
    GPT-4.1 & GPT-4.1 & 93.97 & -1.56 & \underline{97.19} & \textbf{-0.82} & 85.74 & -2.30 & \underline{89.18} & \underline{-1.98} \\
    DeepSeek-V3 & DeepSeek-V3 & 92.51 & -1.80 & 96.92 & -0.91 & 83.08 & -2.66 & 88.07 & -2.22 \\
    
    \specialrule{0em}{1pt}{1pt}
    \hdashline
    \specialrule{0em}{1pt}{1pt}
    \multicolumn{10}{c}{\textbf{\textit{Reasoning-capable models}}} \\
    \specialrule{0em}{1pt}{1pt}
    QwQ-32B & QwQ-32B & 93.23 & -1.69 & 96.41 & -1.00 & 84.23 & -2.48 & 85.81 & -2.66 \\
    DeepSeek-R1 & DeepSeek-R1 & 94.38 & -1.49 & 97.09 & -0.84 & 85.94 & -2.23 & 88.84 & -2.05 \\
    \hline
    \specialrule{0em}{1pt}{1pt}
    \rowcolor{gray!30}
    \multicolumn{10}{c}{\textbf{\textit{Few-shot}}} \\
    \specialrule{0em}{1pt}{1pt}
    \hline
    \specialrule{0em}{1pt}{1pt}

    Qwen2.5-7B & Qwen2.5-7B & 89.04 & -2.66 & 93.70 & -1.60 & 78.22 & -3.51 & 83.93 & -3.17 \\
    Qwen2.5-14B & Qwen2.5-14B & 91.66 & -1.95 & 95.46 & -1.24 & 82.74 & -2.70 & 85.40 & -2.80 \\
    Qwen2.5-32B & Qwen2.5-32B & 91.61 & -1.92 & 95.63 & -1.18 & 82.12 & -2.77 & 86.17 & -2.55 \\
    Qwen2.5-72B & Qwen2.5-72B & 93.20 & -1.74 & 97.03 & -0.87 & 84.26 & -2.52 & 87.33 & -2.26 \\
    GPT-4.1 & GPT-4.1 & 94.17 & -1.50 & 97.06 & \textbf{-0.82} & 86.39 & -2.25 & \underline{89.18} & -2.02 \\
    DeepSeek-V3 & DeepSeek-V3 & 93.15 & -1.68 & 96.81 & -0.92 & 83.94 & -2.53 & 87.70 & -2.24 \\

    \hline
    \specialrule{0em}{1pt}{1pt}
    \rowcolor{gray!30}
    \multicolumn{10}{c}{\textbf{\textit{TACTIC}}} \\
    \specialrule{0em}{1pt}{1pt}
    \hline
    \specialrule{0em}{1pt}{1pt}
    \multicolumn{10}{c}{\textbf{\textit{Non-reasoning models}}} \\
    \specialrule{0em}{1pt}{1pt}

    Qwen2.5-7B & Qwen2.5-7B & 91.43 & -2.00 & 95.96 & -1.09 & 80.39 & -2.93 & 86.55 & -2.53 \\
    Qwen2.5-14B & Qwen2.5-14B & 92.50 & -1.80 & 96.58 & -0.99 & 83.30 & -2.62 & 87.13 & -2.38 \\
    Qwen2.5-32B & Qwen2.5-32B & 93.03 & -1.73 & 96.39 & -0.98 & 83.37 & -2.55 & 87.52 & -2.32 \\
    Qwen2.5-72B & Qwen2.5-72B & 93.09 & -1.69 & 96.79 & -0.91 & 83.94 & -2.54 & 88.02 & -2.23 \\
    DeepSeek-V3 & DeepSeek-V3 & \textbf{95.56} & \underline{-1.43} & \textbf{97.66} & \textbf{-0.82} & \textbf{87.06} & \textbf{-2.07} & \textbf{90.89} & \textbf{-1.97} \\

    \specialrule{0em}{1pt}{1pt}
    \hdashline
    \specialrule{0em}{1pt}{1pt}
    \multicolumn{10}{c}{\textbf{\textit{Reasoning-capable models}}} \\
    \specialrule{0em}{1pt}{1pt}
    Qwen2.5-7B & QwQ-32B & 93.51 & -1.61 & 96.37 & -0.98 & 84.82 & -2.34 & 87.92 & -2.19 \\
    Qwen2.5-14B & QwQ-32B & 93.79 & -1.60 & 96.88 & -0.91 & 85.40 & -2.29 & 88.30 & -2.17 \\
    Qwen2.5-32B & QwQ-32B & 93.84 & -1.59 & 96.51 & -0.93 & 85.34 & -2.30 & 88.30 & -2.15 \\
    Qwen2.5-72B & QwQ-32B & 93.80 & -1.55 & 96.57 & -0.92 & 85.54 & -2.29 & 88.42 & -2.15 \\
    Qwen2.5-14B & DeepSeek-R1 & 94.83 & \underline{-1.42} & 97.02 & -0.85 & 86.56 & \textbf{-2.07} & 88.92 & -2.06 \\
    Qwen2.5-32B & DeepSeek-R1 & \underline{94.90} & \textbf{-1.41} & 97.00 & -0.85 & 86.67 & -2.16 & 88.91 & -2.11 \\
    DeepSeek-V3 & DeepSeek-R1 & 94.84 & -1.44 & 97.04 & -0.86 & \underline{86.77} & -2.10 & 88.95 & -2.03 \\
    \bottomrule
  \end{tabularx}
\end{table}